\documentclass[preprints,a4paper,article,accept,pdftex,moreauthors]{Definitions/mdpi}

\firstpage{1} 
\pubvolume{1}
\issuenum{1}
\articlenumber{0}
\pubyear{2022}
\copyrightyear{2022}
\hreflink{https://doi.org/} %

\usepackage{grffile}

\usepackage{amssymb,amsfonts}
\usepackage{color}
\usepackage{siunitx}
\usepackage{url}
\usepackage[labelformat=simple]{subcaption}

\DeclareCaptionLabelFormat{subcaptionlabel}{\normalfont(\textbf{#2}\normalfont)}
\usepackage{makecell}
\usepackage{adjustbox} %
\usepackage{array}
\usepackage{multirow}
\usepackage{rotating}
\usepackage[english]{babel}
\def\Lum{L}
\def\tref{t_\text{ref}} %
\def\pol{p} %
\def\prtl#1#2{\frac{\partial#1}{\partial#2}}

\def\cE{\mathcal{E}} %
\def\numEvents{N_e} %
\def\numPixels{N_p} %
\def\Warp{\mathbf{W}}
\def\btheta{\boldsymbol{\theta}} %

\def\bx{\mathbf{x}}
\def\bparams{\btheta}
\def\Rot{\mathtt{R}}
\def\br{\mathbf{r}} %

\def\pol{p}
\def\velflow{\mathbf{v}}
\def\angvel{\boldsymbol{\omega}} %

\def\cN{\mathcal{N}} %
\def\bmu{\boldsymbol{\mu}} %

\def\flow{\mathbf{f}}
\def\cD{\mathcal{D}} %
\def\cA{\mathcal{A}} %

\def\lambdadiv{\lambda_\text{div}} %
\def\lambdadef{\lambda_\text{def}} %

\def\IWE{I} %
\def\bzero{\mathbf{0}}
\def\mId{\mathtt{Id}} %
\def\variance{\operatorname{Var}}
\def\mJ{\mathtt{J}}
\def\bvec{\mathbf{b}}
\def\mA{\mathtt{A}}

\definecolor{light-gray}{gray}{0.6}
\newcommand\gframe[1]{{\color{light-gray}\frame{#1}}}

\def\MYTITLE{Event Collapse in Contrast Maximization Frameworks}

\Title{\MYTITLE}

\TitleCitation{\MYTITLE}

\Author{Shintaro Shiba $^{1,2,}$* \orcidA{}, 
Yoshimitsu Aoki $^{1}$ and Guillermo Gallego $^{2,3}$\orcidC{}}
\AuthorNames{Shintaro Shiba, Yoshimitsu Aoki and Guillermo Gallego}

\AuthorCitation{Shiba, S.; Aoki, Y.; Gallego, G.}

\address{%
$^{1}$ \quad Department of Electronics and Electrical Engineering, Faculty of Science and Technology, Keio University, 3-14-1, Hiyoshi, Kohoku-ku, Kanagawa 223-8522, Japan; aoki@elec.keio.ac.jp\\
$^{2}$ \quad Department of Electrical Engineering and Computer Science, Technische Universität Berlin, 10587 Berlin,~Germany; guillermo.gallego@tu-berlin.de\\
$^{3}$ \quad Einstein Center Digital Future and Science of Intelligence Excellence Cluster, 10117 Berlin, Germany}

\corres{Correspondence: sshiba@keio.jp}

\abstract{Contrast maximization (CMax) is a framework that provides state-of-the-art results on several event-based computer vision tasks, such as ego-motion or optical flow estimation. However, it may suffer from a problem called event collapse, which is an undesired solution where events are warped into too few pixels. As prior works have largely ignored the issue or proposed workarounds, it is imperative to analyze this phenomenon in detail. Our work demonstrates event collapse in its simplest form and proposes collapse metrics by using first principles of space--time deformation based on differential geometry and physics. We experimentally show on publicly available datasets that the proposed metrics mitigate event collapse and do not harm well-posed warps. To the best of our knowledge, regularizers based on the proposed metrics are the only effective solution against event collapse in the experimental settings considered, compared with other methods. We hope that this work inspires further research to tackle more complex warp models.
}


\begin{document}

\section{Introduction}
\label{sec:intro}

Event cameras \cite{Delbruck08issle,Suh20iscas,Finateu20isscc} offer potential advantages over standard cameras to tackle difficult scenarios (high speed, high dynamic range, low power).
However, new algorithms are needed to deal with the unconventional type of data they produce (per-pixel asynchronous brightness changes, called events) and unlock their advantages \cite{Gallego20pami}.
Contrast maximization (CMax) is an event processing framework that provides state-of-the-art results on several tasks,
such as rotational motion estimation \cite{Gallego17ral,Kim21ral}, 
feature flow estimation and tracking \cite{Zhu17icra,Zhu17cvpr,Seok20wacv,Stoffregen19cvpr,Dardelet21techrxiv},
ego-motion estimation \cite{Gallego18cvpr,Gallego19cvpr,Peng21pami},
3D reconstruction \cite{Gallego18cvpr,Rebecq18ijcv},
optical flow estimation \cite{Zhu19cvpr,Paredes19pami,Paredes21neurips,Shiba22eccv}, 
motion segmentation \cite{Mitrokhin18iros,Stoffregen19iccv,Zhou21tnnls,Parameshwara21icra,Lu21iros},
guided filtering \cite{Duan21pami},
and image reconstruction \cite{Zhang21arxiv}.

The main idea of CMax and similar event alignment frameworks \cite{Nunes21pami,Gu21iccv} is to find the motion and/or scene parameters that align corresponding events (i.e., events that are triggered by the same scene edge), thus achieving motion compensation. 
The framework simultaneously estimates the motion parameters and the correspondences between events (data association).
However, in some cases CMax optimization converges to an undesired solution where events accumulate into too few pixels, a phenomenon called event collapse (Figure \ref{fig:eyecatcher}). 
Because CMax is at the heart of many state-of-the-art event-based motion estimation methods, 
it is important to understand the above limitation and propose ways to overcome it.
Prior works have largely ignored the issue or proposed workarounds without analyzing the phenomenon in detail.
A more thorough discussion of the phenomenon is overdue, which is the goal of this work.

Contrary to the expectation that event collapse occurs when the event transformation becomes sufficiently complex \cite{Zhu19cvpr,Nunes21pami},
we show that it may occur even in the simplest case of one degree-of-freedom (DOF) motion.
Drawing inspiration from differential geometry and electrostatics, we propose principled metrics to quantify event collapse and discourage it by incorporating penalty terms in the event alignment objective function.
Although event collapse depends on many factors, our strategy aims at modifying the objective's landscape 
to improve the well-posedness of the problem and 
be able to use well-known, standard optimization algorithms.

\def\figWidth{0.225\linewidth}
\def\figWidthLong{0.28\linewidth}

\begin{figure}[H]
    {\scriptsize
    \setlength{\tabcolsep}{2pt}
	\begin{tabular}{
	>{\centering\arraybackslash}m{\figWidthLong}
	>{\centering\arraybackslash}m{\figWidth} 
	>{\centering\arraybackslash}m{\figWidth}
	>{\centering\arraybackslash}m{\figWidth}
	}
		{\includegraphics[clip,trim={0cm 10cm 23cm .3cm},width=\linewidth]{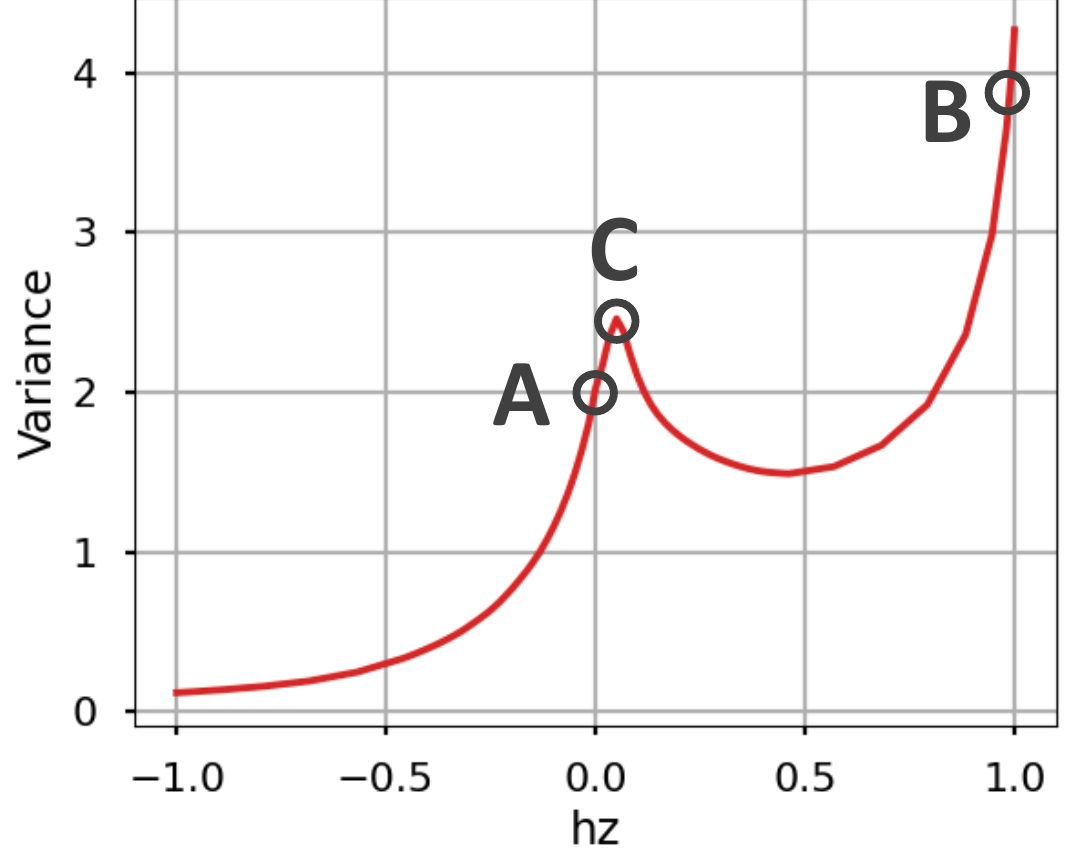}}
		&\gframe{\includegraphics[width=\linewidth]{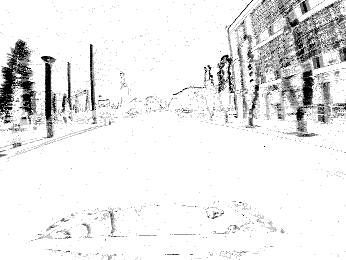}}
		&\gframe{\includegraphics[width=\linewidth]{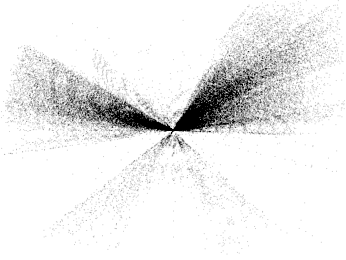}}
		&\gframe{\includegraphics[width=\linewidth]{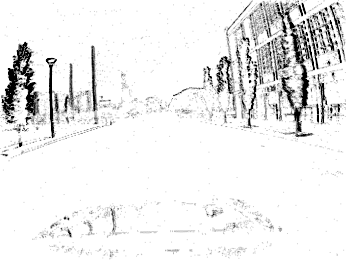}}
		\\

		Loss Landscape
		& Original events at \textbf{A}
		& Collapsed IWE at \textbf{B}
		& Desired IWE at \textbf{C}
		\\
	\end{tabular}
	}
    \caption{\emph{Event~Collapse.}
    \textbf{Left}: Landscape of the image variance loss as a function of the warp parameter $h_z$.
    \textbf{Right}: The IWEs at the different $h_z$ marked in the landspace.
    (\textbf{A}) Original events (identity warp), accumulated over a small $\Delta t$ (polarity is not used).
    (\textbf{B}) Image of warped events (IWE) showing event collapse due to maximization of the objective function.  
    (\textbf{C}) Desired IWE solution using our proposed regularizer: sharper than (\textbf{A}) while avoiding event collapse (\textbf{C}).
}
\label{fig:eyecatcher}
\end{figure}

In summary, our contributions are:
\begin{enumerate}
    \item A study of the event collapse phenomenon in regard to event warping and objective functions (Sections \ref{sec:method:simpleexample} and \ref{sec:experim}).
    \item Two principled metrics of event collapse (one based on flow divergence and one based on area-element deformations)
    and their use as regularizers to mitigate the above-mentioned phenomenon (Sections \ref{sec:method:regularizers} to \ref{sec:method:augmentedobjective}).
    \item Experiments on publicly available datasets that demonstrate, in comparison with other strategies, the effectiveness of the proposed regularizers (Section \ref{sec:experim}).
\end{enumerate}

To the best of our knowledge, this is the first work that focuses on the paramount phenomenon of event collapse, which may arise in state-of-the-art event-alignment methods.
Our experiments show that the proposed metrics 
mitigate event collapse while they do not harm well-posed warps.

\section{Related Work}
\label{sec:related}

\subsection{Contrast Maximization}
\label{sec:related:CMaxInformal}
\begin{figure}[H]
\includegraphics[clip,trim={0cm 9cm 0cm 0cm},width=\linewidth]{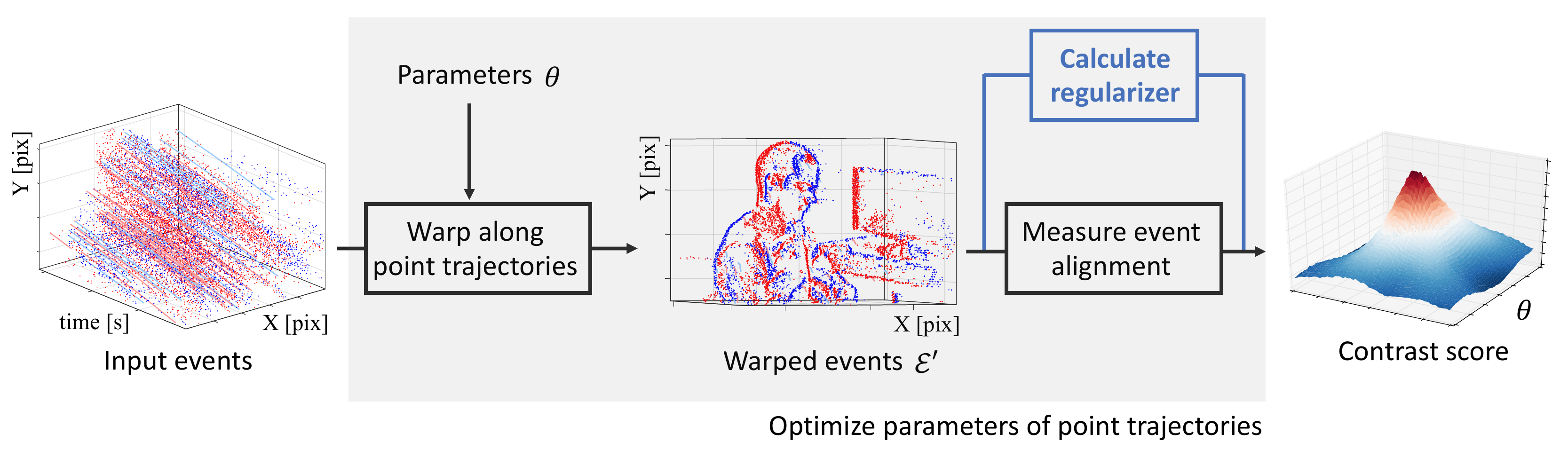}
\caption{Proposed modification of the contrast maximization (CMax) framework in \cite{Gallego18cvpr,Gallego19cvpr} to also account for the degree of regularity (collapsing behavior) of the warp.
Events are colored in red/blue according to their polarity.
Reprinted/adapted with permission from Ref. \cite{Gallego19cvpr}, 2019, Gallego et al.
}
\label{fig:CMax:blockdiagram}
\end{figure}

Our study is based on the CMax framework for event alignment (Figure \ref{fig:CMax:blockdiagram}, bottom branch).
The CMax framework is an iterative method with two main steps per iteration: 
transforming events and computing an objective function from such events.
Assuming constant illumination, events are triggered by moving edges, %
and the goal is to find the transformation/warping parameters $\btheta$ (e.g., motion and scene) 
that achieve motion compensation (i.e., alignment of events triggered at different times and pixels), %
hence revealing the %
edge structure that caused the events.
Standard optimization algorithms (gradient ascent, sampling, etc.) can be used to maximize the event-alignment objective.
Upon convergence, the method provides the best transformation parameters and the transformed events, i.e., motion-compensated image of warped events (IWE).

The first step of the CMax framework transforms events according to a motion or deformation model defined by the task at hand.
For instance, camera rotational motion estimation \cite{Gallego17ral,Liu20cvpr} 
often assumes constant angular velocity ($\btheta\equiv\angvel$) during short time spans, 
hence events are transformed following 3-DOF motion curves %
defined on the image plane by candidate values of $\angvel$.
Feature tracking may assume constant image velocity $\btheta\equiv \velflow$ (2-DOF) \cite{Zhu17icra,Stoffregen17acra}, hence events are transformed following straight lines.

In the second step of CMax, several event-alignment objectives have been proposed to measure the goodness of fit between the events and the model \cite{Gallego19cvpr,Stoffregen19cvpr},
establishing connections between visual contrast, sharpness, and depth-from-focus.
Finally, the choice of iterative optimization algorithm also plays a big role in finding the desired motion-compensation parameters.
First-order methods, such as non-linear conjugate gradient (CG), are a popular choice, trading off accuracy and speed \cite{Gallego18cvpr,Stoffregen19iccv,Zhou21tnnls}. 
Exhaustive search, sampling, or branch-and-bound strategies may be affordable for low-dimensional (DOF) search spaces \cite{Liu20cvpr,Peng21pami}.
As will be presented (Section \ref{sec:method}), our proposal consists of modifying the second step by means of a regularizer (Figure \ref{fig:CMax:blockdiagram}, top branch).

\subsection{Event Collapse}
\label{sec:related:eventcollapse}
\emph{In which estimation problems does event collapse appear?} %
At first look, it may appear that event collapse occurs when the number of DOFs in the warp becomes large enough, i.e., for complex motions.
Event collapse has been reported in homographic motions (8~DOFs) \cite{Nunes21pami,Ozawa22sensors} 
and in dense optical flow estimation \cite{Zhu19cvpr}, where an artificial neural network (ANN) predicts a flow field with $2N_p$ DOFs ($N_p$ pixels), 
whereas it does not occur in feature flow (2~DOFs) or rotational motion flow (3~DOFs). 
However, a more careful analysis reveals that this is not the entire story because event collapse may occur even in the case of 1 DOF, as we show.

\emph{How did previous works tackle event collapse?}
Previous works have tackled the issue in several ways, such as:
(i) initializing the parameters sufficiently close to the desired solution (in the basin of attraction of the local optimum) \cite{Gallego18cvpr}; %
(ii) reformulating the problem, changing the parameter space to reduce the number of DOFs and increase the well-posedness of the problem \cite{Ozawa22sensors,Peng21pami};
(iii) providing additional data, such as depth \cite{Nunes21pami}, thus changing the problem from motion estimation given only events to motion estimation given events and additional sensor data;
(iv) whitening the warped events %
before computing the objective~\cite{Nunes21pami}; and
(v) redesigning the objective function and possibly adding a strong classical regularizer (e.g., Charbonnier loss) \cite{Zhu19cvpr,Stoffregen19cvpr}.
Many of the above mitigation strategies are task-specific because it may not always be possible to consider additional data or reparametrize the estimation problem. 
Our goal is to approach the issue without the need for additional data or changing the parameter space, 
and to show how previous objective functions and newly regularized ones handle event collapse.

\section{Method}
\label{sec:method}
Let us present our approach to measure and mitigate event collapse.
First, we revise how event cameras work (Section \ref{sec:method:eventcamworks}) and the CMax framework (Section \ref{sec:method:cmax}), which was informally introduced in Section \ref{sec:related:CMaxInformal}.
Then, Section \ref{sec:method:simpleexample} builds our intuition on event collapse by analyzing a simple example.
Section \ref{sec:method:regularizers} presents our proposed metrics for event collapse, based on 1-DOF and 2-DOF warps.
Section \ref{sec:method:higherdof} specifies them for higher DOFs,
and Section~\ref{sec:method:augmentedobjective} presents the regularized objective function.

\subsection{How Event Cameras Work}
\label{sec:method:eventcamworks}
Event cameras, such as the Dynamic Vision Sensor (DVS) \cite{Lichtsteiner08ssc,Suh20iscas,Finateu20isscc}, are bio-inspired sensors that capture pixel-wise intensity changes, called events, instead of intensity images. 
An event $e_k \doteq (\bx_k, t_k, \pol_{k})$ is triggered as soon as the logarithmic intensity $\Lum$ at a pixel exceeds a contrast sensitivity $C>0$, 
\begin{equation}
\label{eq:generativeEventCondition}
\Lum(\bx_k,t_k) - \Lum(\bx_k, t_k-\Delta t_k) = \pol_k \, C,
\end{equation} 
where $\bx_k\doteq (x_k, y_k)^{\top}$, $t_k$ (with \si{\micro\second} resolution) and polarity $\pol_{k} \in \{+1,-1\}$
are the spatio-temporal coordinates and sign of the intensity change, respectively,
and $t_k-\Delta t_k$ is the time of the previous event at the same pixel $\bx_k$.
Hence, each pixel has its own sampling rate, which depends on the visual input.

\subsection{Mathematical Description of the CMax Framework}
\label{sec:method:cmax}
The CMax framework \cite{Gallego18cvpr} transforms events in a set $\cE = \{e_k\}_{k=1}^{\numEvents}$ geometrically 
\begin{equation}
e_k \doteq (\bx_k,t_k,\pol_k) \quad\stackrel{\Warp}{\mapsto}\quad
e'_k \doteq (\bx'_k,\tref,\pol_k),
\end{equation}
according to a motion model $\Warp$, producing a set of warped events $\cE' = \{e'_k\}_{k=1}^{\numEvents}$.
The warp $\bx'_k = \Warp(\bx_k,t_k; \bparams)$ transports each event along the point trajectory that passes through it (Figure \ref{fig:CMax:blockdiagram}, left), until $\tref$ is reached.
The point trajectories are parametrized by $\bparams$, which contains the motion and/or scene unknowns.
Then, an objective function~\cite{Gallego19cvpr,Stoffregen19cvpr} measures the alignment of the warped events $\cE'$.
Many objective functions are given in terms of the count of events along the point trajectories, which is called the image of warped events~(IWE):
\begin{equation}
\label{eq:IWE}
\IWE(\bx;\bparams) \doteq \sum_{k=1}^{\numEvents} b_k \,\delta (\bx - \bx'_k(\bparams)).
\end{equation}

Each IWE pixel $\bx$ sums the values of the warped events $\bx'_k$ that fall within it: 
$b_k=\pol_k$ if polarity is used or $b_k=1$ if polarity is not used. %
The Dirac delta $\delta$ is in practice replaced by a smooth approximation~\cite{Ng22ral}, such as a Gaussian, $\delta(\bx-\bmu)\approx\cN(\bx;\bmu,\epsilon^2)$ with $\epsilon=1$~pixel.
A popular objective function $G(\bparams)$ is the visual contrast of the IWE~\eqref{eq:IWE}, given by the variance
\begin{equation}
\label{eq:IWEVariance}
G(\bparams) \equiv \variance\bigl(\IWE(\bx;\bparams)\bigr) 
\doteq \frac{1}{|\Omega|} \int_{\Omega} (\IWE(\bx;\bparams)-\mu_{\IWE})^2 d\bx,
\end{equation}
with mean $\mu_{\IWE} \doteq \frac{1}{|\Omega|} \int_{\Omega} \IWE(\bx;\bparams) d\bx$ and image domain $\Omega$.
Hence, the alignment of the transformed events $\cE'$ (i.e., the candidate ``corresponding events'', triggered by the same scene edge) is measured by the strength of the edges of the IWE.
Finally, an optimization algorithm iterates the above steps until the best parameters are found: 
\begin{equation}
\label{eq:bestParamsOriginal}
\bparams^\ast = \arg\max_{\bparams}G(\bparams).
\end{equation}

\subsection{Simplest Example of Event Collapse: 1 DOF}
\label{sec:method:simpleexample}
To analyze event collapse in the simplest case, 
let us consider an approximation to a translational motion of the camera along its optical axis $Z$ (1-DOF warp).
In theory, translational motions also require the knowledge of the scene depth. 
Here, inspired by the 4-DOF in-plane warp in \cite{Mitrokhin18iros} that approximates a 6-DOF camera motion, we consider a simplified warp that does not require knowledge of the scene depth.
In terms of data, let us consider events from one of the driving sequences of the standard MVSEC dataset \cite{Zhu18ral} (Figure \ref{fig:eyecatcher}).

For further simplicity, let us normalize the timestamps of $\cE$ to the unit interval $t\in [t_1, t_{\numEvents}] \mapsto \tilde{t}\in [0,1]$, and assume a coordinate frame at the center of the image plane, then the warp $\Warp$ is given by
\begin{equation}
\label{eq:warp:hz}
\bx'_k = (1 - \tilde{t}_k h_z)\, \bx_k,
\end{equation}
where $\bparams \equiv h_z$.
Hence, events are transformed along the radial direction from the image center, acting as a virtual focus of expansion (FOE) (cf. the true FOE is given by the data).
Letting the scaling factor in \eqref{eq:warp:hz} be $s_{k} \doteq 1 - \tilde{t}_k h_z$, 
we observe the following:
(i) $s_k$ cannot be negative since it would imply that at least one event has flipped the side on which it lies with respect to the image center;
(ii) if $s_k > 1$ the warped event gets away from the image center (``expansion'' or ``zoom-in''); and
(iii) if $s_k \in [0,1)$ the warped event gets closer to the image center (``contraction'' or ``zoom-out'').
The equivalent conditions in terms of $h_z$ are: (i) $h_z<1$, (ii) $h_z<0$ is an expansion, and (iii) $0<h_z<1$ is a contraction.

\begin{figure}[t]
\centering
\begin{subfigure}{.31\linewidth}
  \centering
  {\includegraphics[clip,trim={1cm 10cm 24cm 2cm},width=\linewidth]{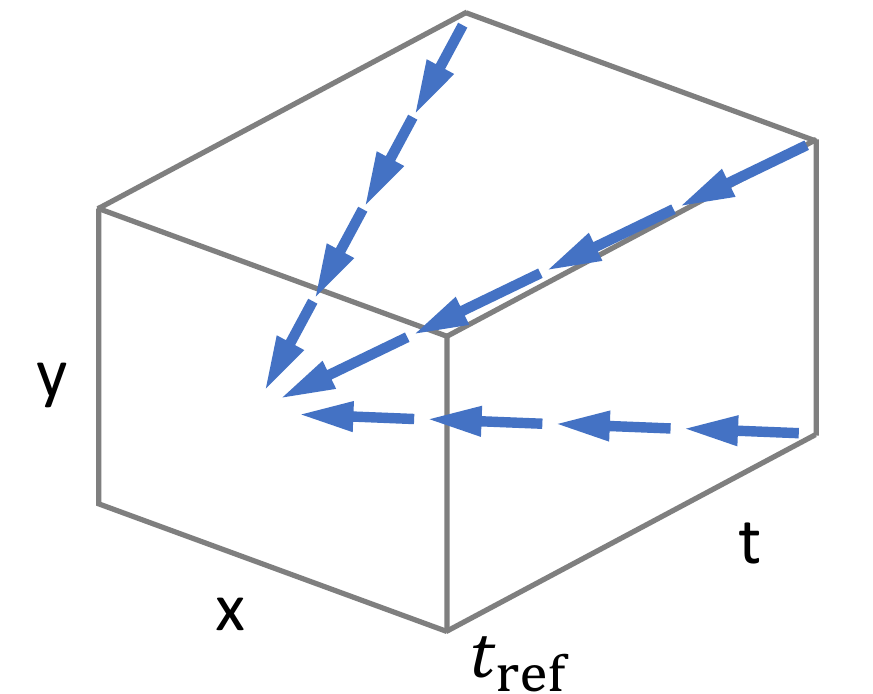}}
  \caption{\centering  }
  \label{fig:spacetime:onedof}
\end{subfigure}\;\;
\begin{subfigure}{.31\linewidth}
  \centering
  {\includegraphics[clip,trim={1cm 10cm 24cm 2cm},width=\linewidth]{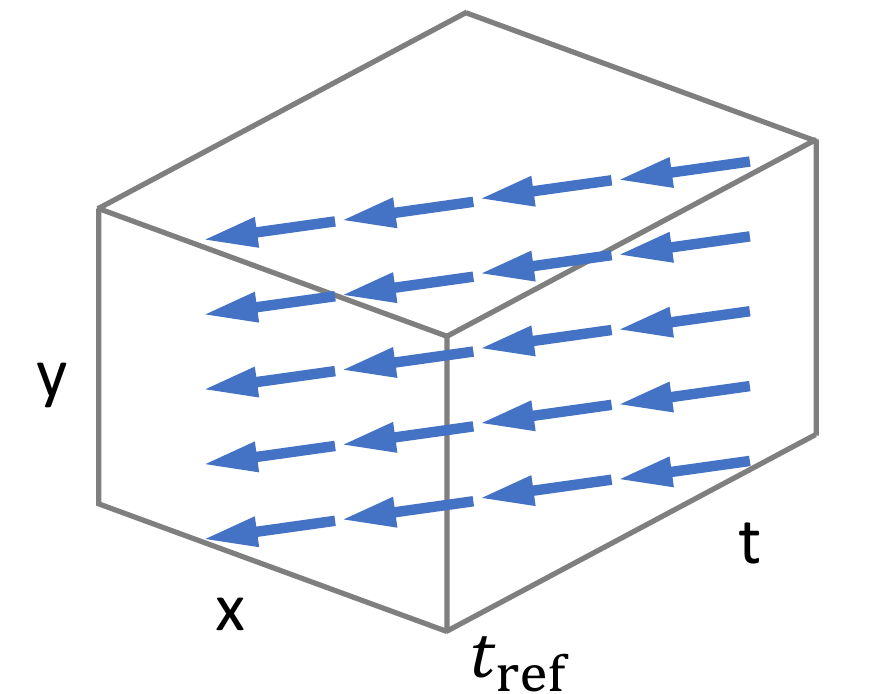}}
  \caption{\centering }
  \label{fig:spacetime:linvel}
\end{subfigure}\;\;
\begin{subfigure}{.31\linewidth}
  \centering
  {\includegraphics[clip,trim={1cm 10cm 24cm 2cm},width=\linewidth]{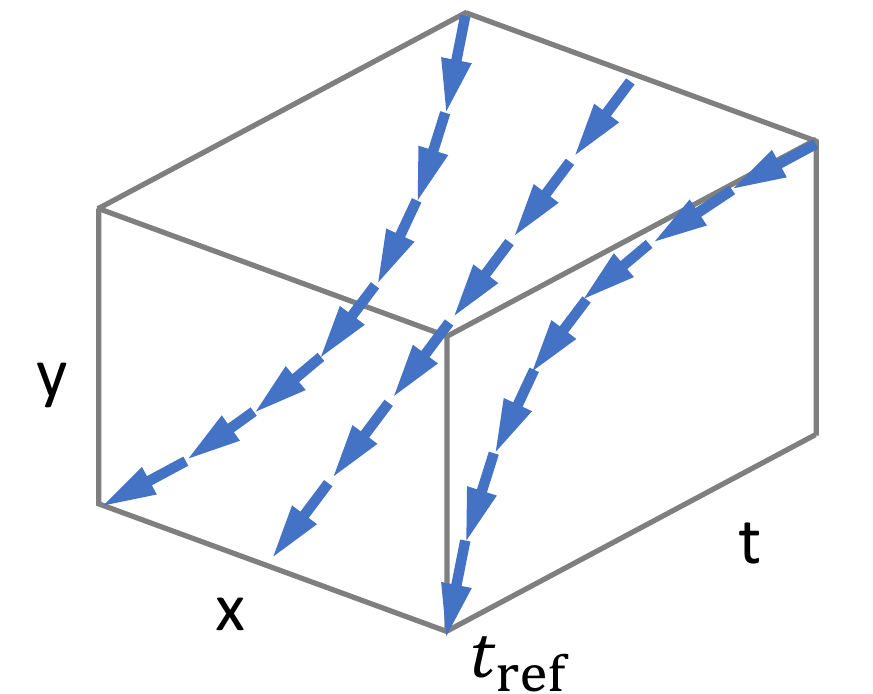}}
  \caption{\centering }
  \label{fig:spacetime:rot}
\end{subfigure}
\\
\caption{\emph{Point trajectories} %
 (streamlines) defined on $x-y-t$ image space by various warps. (\textbf{a}) Zoom in/out warp from image center (1 DOF). (\textbf{b}) Constant image velocity warp (2 DOF). (\textbf{c}) Rotational warp around $X$ axis (3 DOF).
}
\label{fig:spacetime}
\end{figure}

\textls[-45]{Intuitively, event collapse occurs if the contraction is large ($0< s_k \ll 1$) (see Figures \ref{fig:eyecatcher}C and \ref{fig:spacetime:onedof}). }
This phenomenon is not specific of the image variance; 
other objective functions lead to the same result.
As we see, the objective function has a local maximum at the desired motion parameters (Figure \ref{fig:eyecatcher}B). 
The optimization over the entire parameter space converges to a global optimum that explains the event collapse.

\subsubsection*{Discussion}

The above example shows that event collapse is enabled (or disabled) by the type of warp.
If the warp does not enable event collapse 
(contraction or accumulation of flow vectors cannot happen due to the geometric properties of the warp), as in the case of feature flow (2 DOF) \cite{Zhu17icra,Stoffregen17acra} (Figure \ref{fig:spacetime:linvel}) or rotational motion flow (3 DOF) \cite{Gallego17ral,Liu20cvpr} (Figure \ref{fig:spacetime:rot}), then the optimization problem is well posed and multiple objective functions can be designed to achieve event alignment \cite{Gallego19cvpr,Stoffregen19cvpr}. 
However, the disadvantage is that the type of warps that satisfy this condition may not be rich enough to describe complex scene motions.

On the other hand, if the warp allows for event collapse, more complex scenarios can be described by such a broader class of motion hypotheses, but the optimization framework designed for non-event-collapsing scenarios (where the local maximum is assumed to be the global maximum) may not hold anymore. 
Optimizing the objective function may lead to an undesired solution with a larger value than the desired one. 
This depends on multiple elements: the landscape of the objective function (which depends on the data, the warp parametrization, and the shape of the objective function),
and the initialization and search strategy of the optimization algorithm used to explore such a landscape.
The challenge in this situation is to overcome the issue of multiple local maxima and make the problem better posed.
Our approach consists of characterizing event collapse via novel metrics and including them in the objective function as weak constraints (penalties) to yield a better~landscape.

\subsection{Proposed Regularizers}
\label{sec:method:regularizers}

\subsubsection{Divergence of the Event Transformation Flow}
\label{sec:method:divergence}

\begin{figure}[H]
\centering
\includegraphics[width=0.6\linewidth]{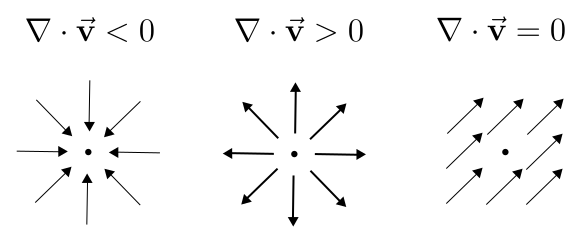}
\caption{\emph{Divergence of different vector fields}, 
$\nabla\cdot \mathbf{v}= \partial_x \mathbf{v}_x + \partial_y \mathbf{v}_y$.
From left to right: contraction (``sink'', leading to event collapse), expansion (``source''), and incompressible fields.
Image adapted from khanacademy.org
}
\label{fig:divergence}
\end{figure}

Inspired by physics, we may think of the flow vectors given by the event transformation $\cE \mapsto \cE'$ as an electrostatic field, 
whose sources and sinks correspond to the location of electric charges %
(Figure \ref{fig:divergence}).
Sources and sinks are mathematically described by the divergence operator $\nabla \cdot\,$. 
Therefore, the divergence of the flow field is a natural choice to characterize event collapse.

The warp $\Warp$ is defined over the space-time coordinates of the events, 
hence its time derivative defines a flow field over space-time:
\begin{equation}
\label{eq:defFlow}
\flow \doteq \prtl{\Warp(\bx,t;\bparams)}{t}.
\end{equation}

For the warp in \eqref{eq:warp:hz}, we obtain $\flow = -h_z \bx$, 
which gives $\nabla \cdot \flow = -h_z \nabla \cdot \bx = -2 h_z$.
Hence, \eqref{eq:warp:hz} defines a constant divergence flow, 
and imposing a penalty on the degree of concentration of the flow field accounts to directly penalizing the value of the parameter $h_z$.

Computing the divergence at each event gives the set 
\begin{equation}
\cD (\cE,\bparams) \doteq \{\nabla \cdot \flow_k\}_{k=1}^{\numEvents}, 
\end{equation}
from which we can compute statistical scores (mean, median, $\min$, etc.):
\begin{align}
    R_D(\cE,\bparams) & \doteq \frac1{\numEvents}\sum_{k=1}^{\numEvents} \nabla \cdot \flow_k. & (\text{mean})
\end{align}

To have a 2D visual representation (``feature map'') of collapse, we build an image (like the IWE) by taking some statistic of the values $\nabla \cdot \flow_k$ that warp to each pixel,
such as the ``average divergence per pixel'':
\begin{equation}
\label{eq:DIWE}
    \text{DIWE}(\bx; \cE,\bparams) \doteq \frac1{\numEvents(\bx)} \sum_{k} (\nabla \cdot \flow_k)\, \delta(\bx - \bx'_k),
\end{equation}
where $\numEvents(\bx)\doteq \sum_{k} \delta(\bx - \bx'_k)$ is the number of warped events at pixel $\bx$ (the IWE).
Then we aggregate further into a score, such as the mean: 
\begin{equation}
R_{\text{DIWE}}(\cE,\bparams) \doteq \frac1{|{\Omega}|} \int_{\Omega} \text{DIWE}(\bx; \cE,\bparams)d\bx.
\end{equation}

In practice we focus on the collapsing part by computing a trimmed mean: 
the mean of the DIWE pixels smaller than a margin $\alpha$ ($-0.2$ in the experiments). 
Such a margin does not penalize small, admissible deformations.

\subsubsection{Area-Based Deformation of the Event Transformation}
\label{sec:method:deformation}

\def\figWidth{0.31\linewidth}

\begin{figure}[H]
    {\scriptsize
    \setlength{\tabcolsep}{5pt}
	\begin{tabular}{
	>{\centering\arraybackslash}m{\figWidth} 
	>{\centering\arraybackslash}m{\figWidth}
	>{\centering\arraybackslash}m{\figWidth}
	}
		{\includegraphics[clip,trim={1cm 8.5cm 21.5cm 1cm},width=\linewidth]{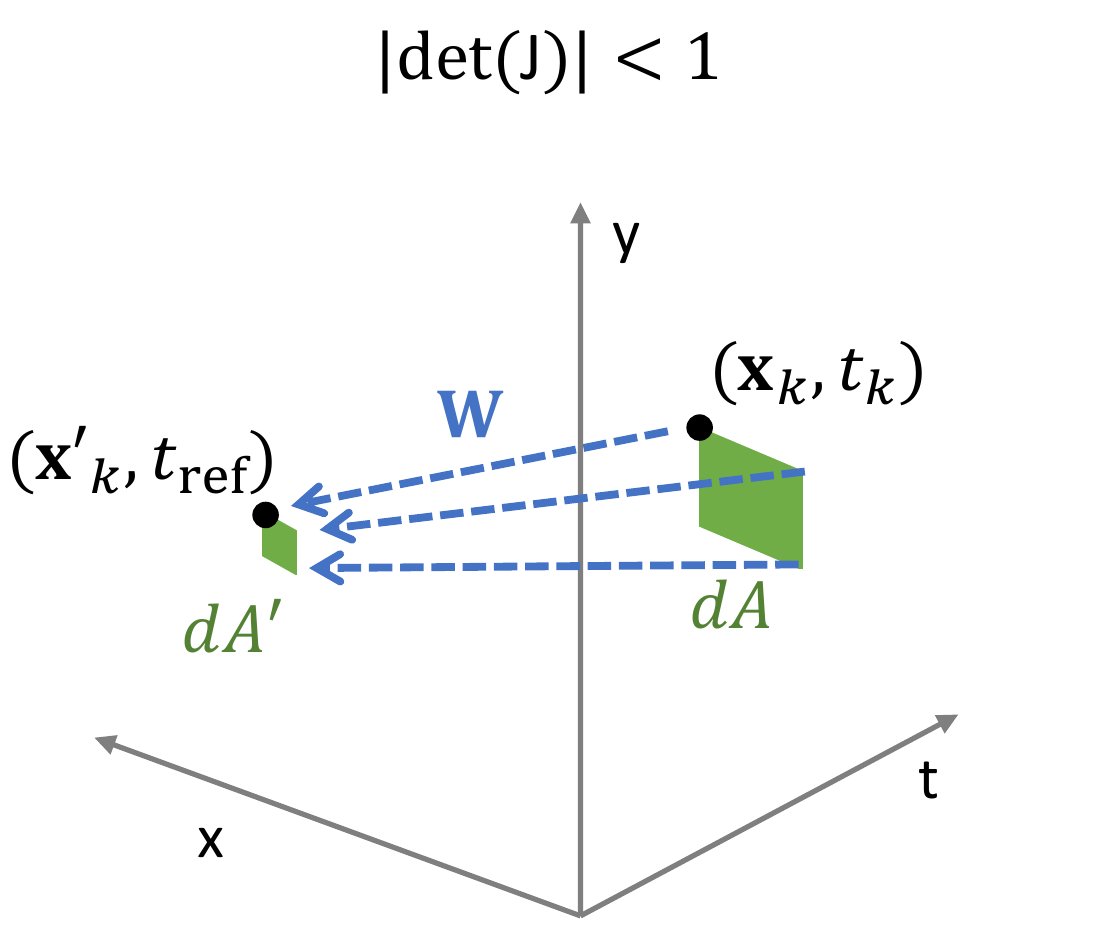}}
		&{\includegraphics[clip,trim={1cm 8.5cm 21.5cm 1cm},width=\linewidth]{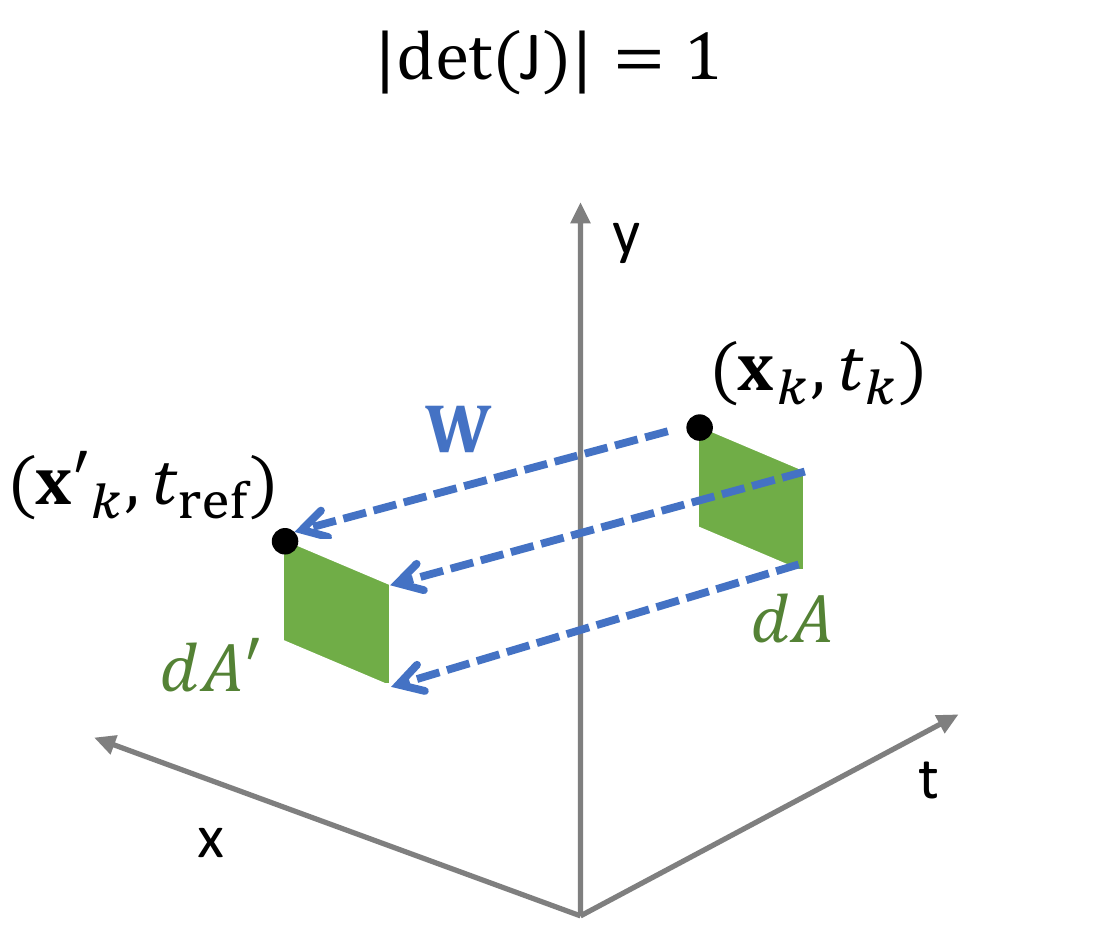}}
		&{\includegraphics[clip,trim={.6cm 8.5cm 21.5cm 1cm},width=\linewidth]{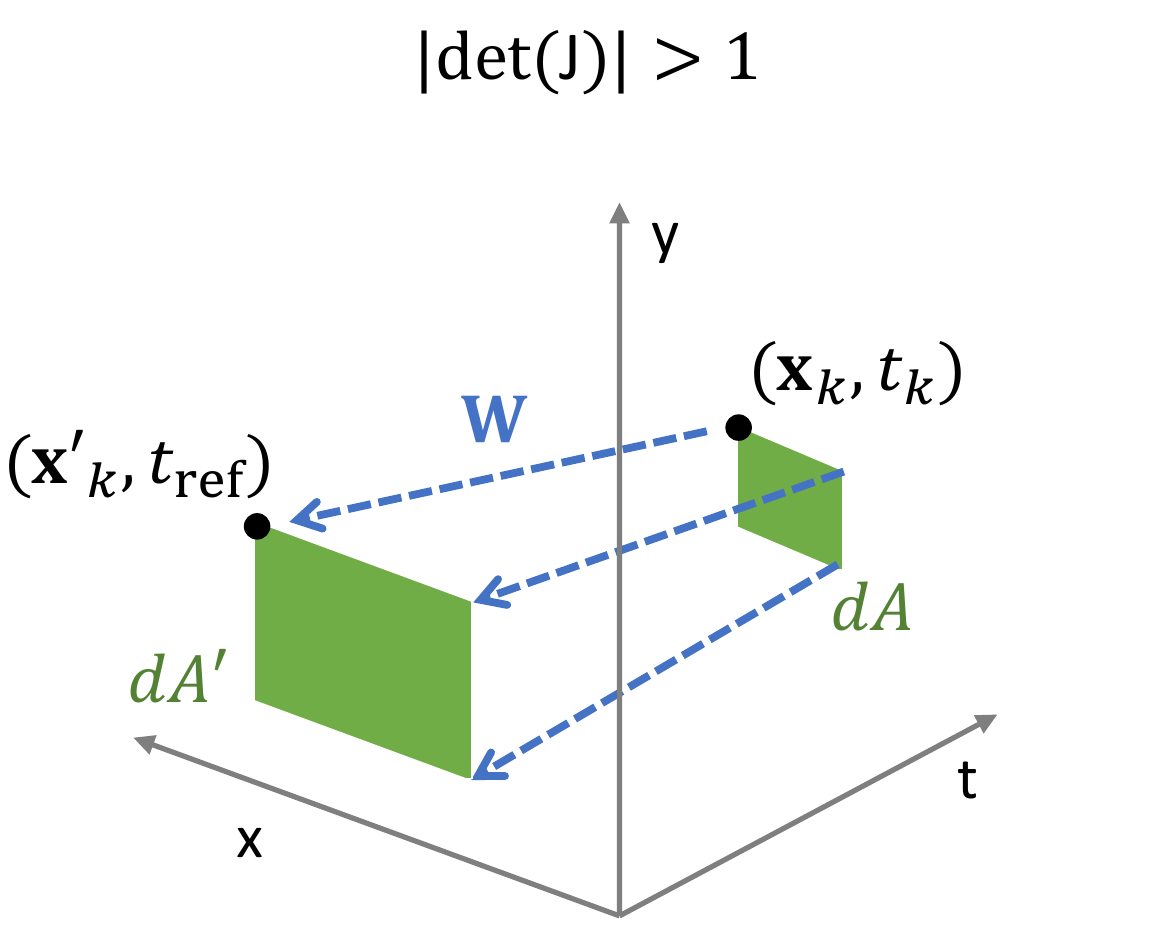}}
		\\
		Contraction
		& No change of area
		& Expansion\\
	\end{tabular}
	}
    \caption{
    \emph{Area deformation of various warps}.
    An area of $dA~\text{pix}^2$ at $(\bx_k,t_k)$ and is warped to $\tref$, 
    giving an area $dA' = |\det(\mJ_k)| dA~\text{pix}^2$ at $(\bx'_k,\tref)$,
    where $\mJ_k \equiv \mJ(e_k) \equiv \mJ(\bx_k,t_k;\bparams)$ (see \eqref{eq:defJacobian}).
    From left to right, increasing area amplification factor $|\det(\mJ)| \in [0, \infty)$.
}
\label{fig:deformation}
\end{figure}

In addition to vector calculus, %
we may also use tools from differential geometry to characterize event collapse.
Building on~\cite{Gallego18cvpr}, the point trajectories define the streamlines of the transformation flow, and we may measure how they concentrate or disperse based on how the area element deforms along them. 
That is, we consider a small area element $dA=dxdy$ attached to each point along the trajectory 
and measure how much it deforms when transported to the reference time:
$dA' = |\det(\mJ)|\, dA$, with the Jacobian 
\begin{equation}
\label{eq:defJacobian}
\mJ(\bx,t;\bparams) \doteq \prtl{\Warp(\bx,t;\bparams)}{\bx}
\end{equation}
(see Section \ref{fig:deformation}). 
The determinant of the Jacobian is the amplification factor: $|\det(\mJ)|>1$ if the area expands, and $|\det(\mJ)|<1$ if the area shrinks.

For the warp in \eqref{eq:warp:hz}, we have the Jacobian 
$\mJ %
= (1 - \tilde{t} h_z)\mId$, and so $\det(\mJ) = (1 - \tilde{t} h_z)^2$.
Interestingly, the area deformation around event $e_k$, 
$\mJ(e_k) \equiv \mJ(\bx_k,t_k;\bparams)$, is directly related to the scaling factor $s_k$:
$\det(\mJ(e_k)) = s_k^2$.

Computing the amplification factors at each event gives the set
\begin{equation}
\cA (\cE,\bparams) \doteq \bigl\{|\det(\mJ(e_k))|\bigr\}_{k=1}^{\numEvents},
\end{equation}
from which we can compute statistical scores.
For example, 
\begin{align}
    R_A(\cE,\bparams) & \doteq \frac1{\numEvents}\sum_{k=1}^{\numEvents} |\det(\mJ(e_k))| & (\text{mean})
\end{align}
gives an average score: $R_A>1$ for expansion, and $R_A<1$ for contraction.

\textls[-15]{We build a deformation map (or image of warped areas (IWA)) 
by taking some statistic of the values $|\det(\mJ(e_k))|$ that warp to each pixel,
such as the ``average amplification per pixel'':}
\begin{equation}
\label{eq:iwa}
\text{IWA}(\bx) \doteq 1 + \frac1{\numEvents(\bx)} \sum_{k=1}^{\numEvents} \bigl(|\det(\mJ(e_k))| - 1\bigr) \,\delta(\bx - \bx'_k).
\end{equation}

This assumes that if no events warp to a pixel $\bx_p$, then $N_e(\bx_p)=0$, and there is no deformation ($\text{IWA}(\bx_p) = 1$).
Then, we summarize the deformation map into a score, such as the mean:
\begin{equation}
R_{\text{IWA}}(\cE,\bparams) \doteq \frac1{|{\Omega}|} \int_{\Omega} \text{IWA}(\bx; \cE,\bparams) d\bx.
\end{equation}

To concentrate on the collapsing part, we compute a trimmed mean: 
the mean of the IWA pixels smaller than a margin $\alpha$ ($0.8$ in the experiments). 
The margin approves small, admissible deformations.

\subsection{Higher DOF Warp Models}
\label{sec:method:higherdof}
\subsubsection{Feature Flow}
\label{sec:method:featureflow}
Event-based feature tracking is often described by the warp
$\Warp(\bx,t;\bparams) = \bx + (t -t_\text{ref}) \bparams$, which assumes constant image velocity $\bparams$ (2 DOFs) over short time intervals.
As expected, the flow for this warp coincides with the image velocity, $\flow = \bparams$, which is independent of the space-time coordinates ($\bx,t$).
Hence, the flow is incompressible ($\nabla \cdot \flow = \bzero$): 
the streamlines given by the feature flow do not concentrate or disperse; they are parallel.
Regarding the area deformation, the Jacobian $\mJ = \partial (\bx + (t-t_\text{ref})\bparams) / \partial \bx = \mId$ is the identity matrix.
Hence $|\det(\mJ)|=1$, that is, translations on the image plane do not change the area of the pixels around a point.

In-plane translation warps, such as the above 2-DOF warp, are well-posed and serve as reference to design the regularizers that measure event collapse. 
It is sensible for well-designed regularizers to penalize warps whose characteristics deviate from those of the reference warp: 
zero divergence and unit area amplification factor.

\subsubsection{Rotational Motion}
\label{sec:method:rotationalmotion}
As the previous sections show, the proposed metrics designed for the zoom in/out warp produce the expected characterization of the 2-DOF feature flow (zero divergence and unit area amplification), which is a well-posed warp.
Hence, if they were added as penalties into the objective function they would not modify the energy landscape.
We now consider their influence on rotational motions, which are also well-posed warps.
In particular, we consider the problem of estimating the angular velocity of a predominantly rotating event camera by means of CMax, which is a popular research topic \cite{Gallego17ral,Liu20cvpr,Peng21pami,Nunes21pami,Gu21iccv}.
By using calibrated and homogeneous coordinates, the warp is given by 
\begin{equation}
\label{eq:warp:rotangvel}
\bx^{h\prime} \sim \Rot(t \angvel)\,\bx^h,
\end{equation}
where $\bparams \equiv \angvel = (\omega_{1},\omega_{2},\omega_{3})^{\top}$ is the angular velocity, 
$t\in [0,\Delta t]$,
and $\Rot$ is parametrized by using exponential coordinates (Rodrigues rotation formula \cite{Murray94book,Gallego14jmiv}).

Divergence:
It is well known that the flow is $\flow = B(\bx)\,\angvel$, 
where $B(\bx)$ is the rotational part of the feature sensitivity matrix \cite{Corke17book}. 
Hence
\begin{equation}
\label{eq:rot:flowdivergence}
\nabla \cdot \flow = 3 (x\omega_2 - y\omega_1).
\end{equation}

Area element: 
Letting $\br_3^\top$ be the third row of $\Rot$, and using (32)--(34) in~\cite{Gallego11tgrs},
\begin{equation}
\label{eq:rot:detJacArea}
\det(\mJ) = (\br_3^\top \bx^h)^{-3}.
\end{equation}

Rotations around the $Z$ axis clearly present no deformation, 
regardless of the amount of rotation, and this is captured by the proposed metrics because: 
(i) the divergence is zero, thus the flow is incompressible,
and (ii) $\det(\mJ)=1$ since $\br_3=(0,0,1)^\top$ and $\bx^h=(x,y,1)^\top$.
For other, arbitrary rotations, there are deformations, but these are mild if the rotation angle $\Delta t \|\angvel\|$ is small.

\subsubsection{Planar Motion}
Planar motion is the term used to describe the motion of a ground robot that can translate and rotate freely on a flat ground.
If such a robot is equipped with a camera pointing upwards or downwards, the resulting motion induced on the image plane, parallel to the ground plane, is an isometry (Euclidean transformation).
This motion model is a subset of the parametric ones in \cite{Gallego18cvpr}, and it has been used for CMax in \cite{Peng21pami,Nunes21pami}.
For short time intervals, planar motion may be parametrized by 3 DOFs: linear velocity (2 DOFs) and angular velocity (1 DOF).
As the divergence and area metrics show in the Appendix, planar motion is a well-posed warp. 
The resulting motion curves on the image plane do not lead to event collapse.

\subsubsection{Similarity Transformation}
The 1-DOF zoom in/out warp in Section \ref{sec:method:simpleexample} is a particular case of the 4-DOF warp in~\cite{Mitrokhin18iros},
which is an in-plane approximation to the motion induced by a freely moving camera.
The same idea of combining translation, rotation, and scaling for CMax is expressed by the similarity transformation in \cite{Nunes21pami}.
Both 4-DOF warps enable event collapse because they allow for zoom-out motion curves.
Formulas justifying it are given in the Appendix.

\subsection{Augmented Objective Function}
\label{sec:method:augmentedobjective}
We propose to augment previous objective functions (e.g., \eqref{eq:bestParamsOriginal}) with penalties obtained from the metrics developed above for event collapse:
\begin{equation}
\label{eq:compositeObjective}
\bparams^\ast = \arg\!\min_{\bparams} J(\bparams) = \arg\!\min_{\bparams} \left(-G(\bparams) + \lambda R(\bparams)\right).
\end{equation}

We may interpret $G(\bparams)$ (e.g., contrast or focus score \cite{Gallego19cvpr}) as the data fidelity term and $R(\bparams)$ as the regularizer,
or, in Bayesian terms, the likelihood and the prior, respectively.

\section{Experiments}
\label{sec:experim}

We evaluate our method on publicly available datasets, whose details are described in Section \ref{sec:experim:datasets}.
First, Section \ref{sec:experim:collapse} shows that the proposed regularizers mitigate the overfitting issue on warps that enable collapse.
For this purpose we use driving datasets (MVSEC \cite{Zhu18ral}, DSEC \cite{Gehrig21ral}).
Next, Section \ref{sec:experim:rotation} shows that the regularizers do not harm well-posed warps.
To this end, we use the ECD dataset \cite{Mueggler17ijrr}.
Finally, Section \ref{sec:experim:sensitivity} conducts a sensitivity analysis of the regularizers.

\subsection{Evaluation Datasets and Metrics}
\label{sec:experim:datasets}

\subsubsection{Datasets} 
The \emph{MVSEC} dataset \cite{Zhu18ral} is a widely used dataset for various vision tasks, such as optical flow estimation \cite{Zhu19cvpr,Gehrig21threedv,Nagata21sensors,Paredes21neurips,Shiba22eccv}.
Its sequences are recorded on a drone (indoors) or on a car (outdoors), 
and comprise events, grayscale frames and IMU data from an mDAVIS346 \cite{Taverni18tcsii} ($346 \times 260$ pixels), as well as camera poses and LiDAR data.
Ground truth optical flow is computed as the motion field \cite{Zhu18rss}, given the camera velocity and the depth of the scene (from the LiDAR).
We select several excerpts from the \emph{outdoor\_day1} sequence with a forward motion. This motion is reasonably well approximated by collapse-enabled warps such as \eqref{eq:warp:hz}. 
In total, we evaluate 3.2 million events spanning 10 s.

The \emph{DSEC} dataset \cite{Gehrig21ral} is a more recent driving dataset with a higher resolution event camera (Prophesee Gen3, $640 \times 480$ pixels).
Ground truth optical flow is also computed as the motion field using the scene depth from a LiDAR \cite{Gehrig21threedv}.
We evaluate on the \emph{zurich\_city\_11} sequence, using in total 380 million events spanning 40 s.

\textls[-45]{The \emph{ECD} dataset \cite{Mueggler17ijrr} is the de facto standard to assess event camera ego-motion~\cite{Gallego17ral,Zhu17cvpr,Rosinol18ral,Gu21iccv,Rebecq17ral,Mueggler18tro,Zhou20tro}.}
Each sequence provides events, frames, a calibration file, and IMU data (at 1kHz) from a DAVIS240C camera \cite{Brandli14ssc} ($240 \times 180$ pixels), as well as ground-truth camera poses from a motion-capture system (at 200Hz).
For rotational motion estimation (3DOF), we use the natural-looking \emph{boxes\_rotation} and \emph{dynamic\_rotation} sequences.
We evaluate 43 million events (10 s) of the box sequence, and 15 million events (11 s) of the dynamic sequence.

The driving datasets (MVSEC, DSEC) and the selected sequences in the ECD dataset have different type of motions: 
forward (which enables event collapse) vs.~rotational (which does not suffer from event collapse). 
Each sequence serves a different test purpose, as discussed in the next sections.

\subsubsection{Metrics}
The metrics used to assess optical flow accuracy (MVSEC and DSEC datasets) are 
the average endpoint error (AEE) and the percentage of pixels with AEE greater than $N$ pixels (denoted by ``$N$PE'', for $N=\{3,10,20\}$).
Both are measured over pixels with valid ground-truth values.
We also use the FWL metric \cite{Stoffregen20eccv} to assess event alignment by means of the IWE sharpness (the FWL is the IWE variance relative to that of the identity warp).

Following previous works \cite{Gallego19cvpr,Nunes21pami,Gu21iccv}, rotational motion accuracy is assessed as the RMS error of angular velocity estimation.
Angular velocity $\angvel$ is assumed to be constant over a window of events, estimated and compared with the ground truth at the midpoint of the window.
Additionally, we use the FWL metric to gauge event alignment \cite{Stoffregen20eccv}.

\textls[-5]{The event time windows are as follows: 
the events in the time spanned by $dt=4$~frames} in MVSEC (standard in \cite{Zhu19cvpr,Gehrig21threedv,Paredes21neurips}), 
500k events for DSEC, and 30k events for ECD~\cite{Gu21iccv}.
The regularizer weights for divergence ($\lambdadiv$) and deformation ($\lambdadef$) are as follows:
$\lambdadiv = 2$ and $\lambdadef = 5$ for MVSEC, $\lambdadiv = 50$ and $\lambdadef = 100$ for DSEC, and $\lambdadiv = 5$ and $\lambdadef = 10$ for ECD experiments.

\subsection{Effect of the Regularizers on Collapse-Enabled Warps}
\label{sec:experim:collapse}

Tables \ref{tab:main_mvsec} and \ref{tab:main_dsec} report the results on the MVSEC and DSEC benchmarks, respectively, by
using two different loss functions $G$: 
the IWE variance \eqref{eq:IWEVariance} and the squared magnitude of the IWE gradient, abbreviated ``Gradient Magnitude'' \cite{Gallego19cvpr}.
For MVSEC, we report the accuracy within the time interval of $dt=4$ grayscale frame (at $\approx$ 45Hz).
The optimization algorithm is the Tree-Structured Parzen Estimator (TPE) sampler \cite{Bergstra11nips} for both experiments, with a number of sampling points equal to 300 (1 DOF) and 600 (4 DOF).
The tables quantitatively capture the collapse phenomenon suffered by the original CMax framework \cite{Gallego18cvpr} and the whitening technique \cite{Nunes21pami}.
Their high FWL values indicate that contrast is maximized; however, the AEE and $N$PE values are exceedingly high (e.g., $>80$ pixels, $20\text{PE}>80$\%), indicating that the estimated flow is unrealistic.
\begin{table}[H]
\centering
\caption{Results of MVSEC dataset \cite{Zhu18rss}.
}
\label{tab:main_mvsec}
\adjustbox{max width=\textwidth}{%
\setlength{\tabcolsep}{2pt}
\begin{tabular}{ll*{10}{S[table-format=3.4]}}
\toprule
 &  
 & \multicolumn{5}{c}{\textbf{Variance}}
 & \multicolumn{5}{c}{\textbf{Gradient Magnitude}}
 \\
 \cmidrule(l{1mm}r{1mm}){3-12}
& 
&\textbf{AEE} $\boldsymbol{\downarrow}$ & \text{\textbf{3PE} $\boldsymbol{\downarrow}$} & \text{\textbf{10PE} $\boldsymbol{\downarrow}$} & \text{\textbf{20PE} $\boldsymbol{\downarrow}$} & \text{\textbf{FWL} $\boldsymbol{\uparrow}$}
&\text{\textbf{AEE} $\boldsymbol{\downarrow}$} & \text{\textbf{3PE} $\boldsymbol{\downarrow}$} & \text{\textbf{10PE} $\boldsymbol{\downarrow}$} & \text{\textbf{20PE} $\boldsymbol{\downarrow}$} & \text{\textbf{FWL} $\boldsymbol{\uparrow}$}
\\
\midrule 
 & Ground truth flow & \_ & \_ & \_ & \_ & 1.047993099 & \_ & \_ & \_ & \_ & 1.047993099 \\
 & Identity warp & 4.845102657 & 60.58545033 & 10.38418053 & 0.314555551 & 1.0 &  4.845102657 & 60.58545033 & 10.38418053 & 0.314555551 & 1.0 \\
\midrule

\multirow{5}{*}{\begin{turn}{90}
1 DOF
\end{turn}} 
 & No regularizer & 89.33976 & 97.29911 & 95.41734707 & 92.39056633 & 1.9043804 & 85.774515 & 93.96433 & 86.23523074 & 83.44562201 & 1.8692 \\
 & Whitening \cite{Nunes21pami} & 89.5848 & 97.18128 & 96.77280241 & 93.76030044 & 1.9013 & 81.10176919 & 90.86011026 & 89.0438589 & 86.20025493 & 1.845830887 \\
 & Divergence (Ours) & 3.997768 & 46.01921 & 2.771386599 & 0.05492249224 & 1.119496 & 2.87221497 & 32.67842711 & 2.516948313 & 0.03158205548 & 1.173016021 \\
 & Deformation (Ours) & 4.46584 & 52.60007 & 5.162230828 & 0.127030408 & 1.0766 & 3.97013 & 48.7886 & 3.209162316 & 0.06638392314 & 1.0873 \\
 & Div. + Def. (Ours) & 3.299381 & 33.08753545 & 2.613900583 & 0.4842627055 & 1.196874476 & 2.849103044 & 32.33705658 & 2.435868209 & 0.03246428393 & 1.174927461 \\

\midrule

\multirow{5}{*}{\begin{turn}{90}
4 DOF~~\cite{Mitrokhin18iros}
\end{turn}} 
 & No regularizer & 90.2153977 & 90.2153977 & 96.9406752 & 93.86453651 & 2.046721054 & 91.26328556 & 99.48531996 & 95.06164699 & 91.46305309 & 2.013541325 \\
 & Whitening \cite{Nunes21pami} & 90.81840915 & 99.1098107 & 98.03757781 & 95.0418796 & 2.039896952 & 88.38464766 & 98.86976557 & 92.41199318 & 88.66373937 & 1.999155093 \\
 & Divergence  (Ours) & 7.254550846 & 81.74741226 & 18.53133953 & 0.6921496743 & 1.086032753 & 5.365312689 & 66.17947036 & 10.80987321 & 0.2764757032 & 1.14390564 \\
 & Deformation (Ours) & 8.132825707 & 87.4592456 & 18.53133953 & 1.092239517 & 1.027891481 & 5.253162047  & 64.78740194 & 13.17500467 & 0.3739068138 & 1.151436025 \\
 & Div. + Def. (Ours) & 5.14482574 & 65.60746589 & 10.75469645 & 0.3817697962 & 1.157302645 & 5.409725124 & 66.01420062 & 13.18880475 & 0.5417801136 & 1.14339947 \\

\bottomrule
\end{tabular}
}
\end{table}

\begin{table}[H]
\centering
\caption{Results of DSEC dataset \cite{Gehrig21ral}.
}
\label{tab:main_dsec}
\adjustbox{max width=\textwidth}{%
\setlength{\tabcolsep}{2pt}
\begin{tabular}{ll*{10}{S[table-format=3.4]}}
\toprule
 &  
 & \multicolumn{5}{c}{\textbf{Variance}}
 & \multicolumn{5}{c}{\textbf{Gradient Magnitude}}
 \\
 \cmidrule(l{1mm}r{1mm}){3-12}
& 
&\text{\textbf{AEE} $\boldsymbol{\downarrow}$} & \text{\textbf{3PE} $\boldsymbol{\downarrow}$} & \text{\textbf{10PE} $\boldsymbol{\downarrow}$} & \text{\textbf{20PE} $\boldsymbol{\downarrow}$} & \text{\textbf{FWL} $\boldsymbol{\uparrow}$}
&\text{\textbf{AEE} $\boldsymbol{\downarrow}$} & \text{3\textbf{PE} $\boldsymbol{\downarrow}$} & \text{\textbf{10PE} $\boldsymbol{\downarrow}$} & \text{\textbf{20PE} $\boldsymbol{\downarrow}$} & \text{\textbf{FWL} $\boldsymbol{\uparrow}$}
\\
\midrule 
 & Ground truth flow & \_ & \_ & \_ & \_ & 1.090988475 & \_ & \_ & \_ & \_ & 1.090988475 \\
 & Identity warp & 5.843080683 & 60.45375986 & 16.6455773 & 3.395154843 & 1.0 & 5.843080683 & 60.45375986 & 16.6455773 & 3.395154843 & 1.0 \\
\midrule

\multirow{5}{*}{\begin{turn}{90}
1 DOF
\end{turn}} 
 & No regularizer & 156.1258394 & 99.88354101 & 99.33309776 & 98.18140673 & 2.580704274 & 156.0779055 & 99.92758934 & 99.40158847 & 98.10842233 & 2.579711796 \\
 & Whitening \cite{Nunes21pami} & 156.1784653 & 99.95286001 & 99.51452798 & 98.26119902 & 2.582950586 & 156.8175223 & 99.87981264 & 99.38352644 & 98.32937688 & 2.58066318 \\
 & Divergence (Ours) & 12.48664938 & 69.86051446 & 20.77921847 & 6.664941639 & 1.425272606 & 5.472864325 & 63.48196278 & 14.66048036 & 1.348473519 & 1.344124739 \\
 & Deformation (Ours) & 9.007609466 & 68.95608551 & 18.86118538 & 4.773455209 & 1.402945911 & 5.793269017 & 64.02245664 & 16.10840844 & 2.751044423 & 1.361735211 \\
 & Div. + Def. (Ours) & 6.061261761 & 68.47937073 & 17.08434678 & 2.272174046 & 1.356512051 & 5.526156883 & 64.09306651 & 15.06454271 & 1.368051902 & 1.348987971 \\ 

\midrule

\multirow{5}{*}{\begin{turn}{90}
4 DOF~~\cite{Mitrokhin18iros}
\end{turn}} 
 & No regularizer & 157.542172 & 99.96821514 & 99.64288796 & 98.66698569 & 2.637262171 & 157.3356017 & 99.94254934 & 99.52521317 & 98.44449633 & 2.618987077 \\
 & Whitening \cite{Nunes21pami} & 157.7329839 & 99.9690844 & 99.66131966 & 98.71335243 & 2.604664639 & 156.1180868 & 99.91218462 & 99.25821152 & 97.92947005 & 2.609568805 \\
 & Divergence (Ours) & 14.34541785 & 90.83596808 & 41.61860743 & 10.81636673 & 1.349181226 & 10.42823007 & 91.37707859 & 41.62869772 & 9.433569862 & 1.214208323 \\
 & Deformation (Ours) & 15.11846767 & 94.95566648 & 62.58570184 & 22.6160355 & 1.24728993 & 10.006316 & 90.14615144 & 39.45149429 & 8.670040138 & 1.249295937 \\
 & Div. + Def. (Ours) & 10.05902513 & 90.64687198 & 40.60743178 & 8.583359441 & 1.263127072 & 10.39335383 & 91.01817547 & 41.81212268 & 9.396431802 & 1.226296549 \\

\bottomrule
\end{tabular}
}
\end{table}

\def\figWidth{0.1712\linewidth} \def\figWidthLong{0.214\linewidth} %
\begin{figure}[H]
	\centering
    {\scriptsize
    \setlength{\tabcolsep}{1pt}
	\begin{tabular}{
	>{\centering\arraybackslash}m{0.3cm}
	>{\centering\arraybackslash}m{\figWidth} 
	>{\centering\arraybackslash}m{\figWidth}
	>{\centering\arraybackslash}m{\figWidth}
	>{\centering\arraybackslash}m{\figWidthLong}
	>{\centering\arraybackslash}m{\figWidthLong}
	}
		& Original events
		& IWE w/o regularizer
		& IWE with regularizer
		& Divergence map\;\;\;\;\;
		& Deformation map\;\;\;\;\\
		
		\multirow{2}{*}{\rotatebox{90}{\makecell{MVSEC~~\cite{Zhu18ral}}}}
		&\gframe{\includegraphics[width=\linewidth]{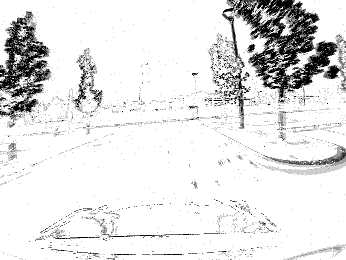}}
		&\gframe{\includegraphics[width=\linewidth]{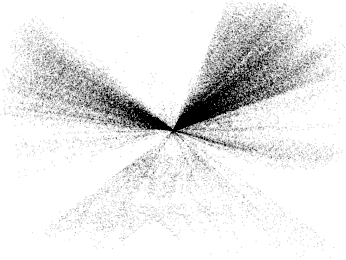}}
		&\gframe{\includegraphics[width=\linewidth]{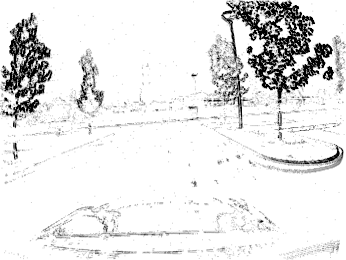}}
		&\includegraphics[clip,trim={.3cm 1.1cm .3cm 1cm},width=\linewidth]{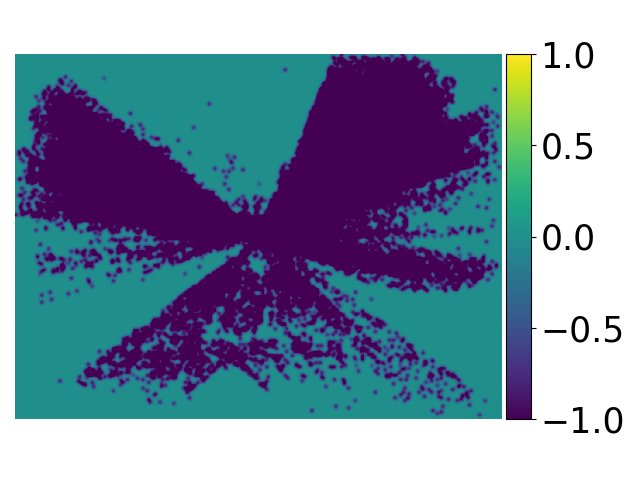}
		&\includegraphics[clip,trim={.3cm .9cm .3cm .8cm},width=\linewidth]{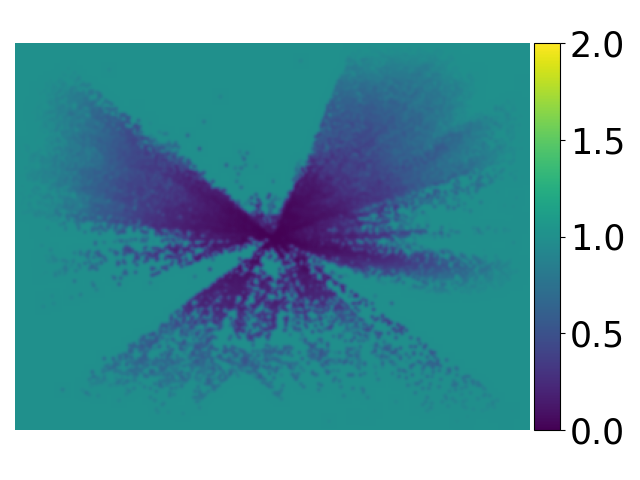}
		\\

		&\gframe{\includegraphics[width=\linewidth]{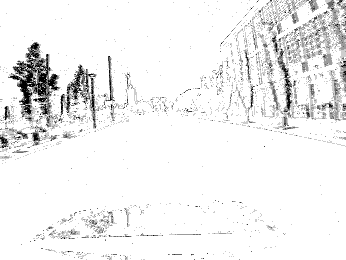}}
		&\gframe{\includegraphics[width=\linewidth]{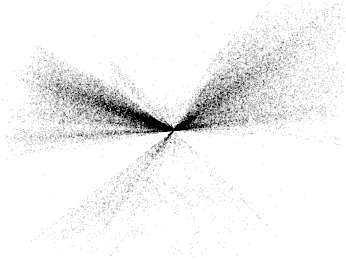}}
		&\gframe{\includegraphics[width=\linewidth]{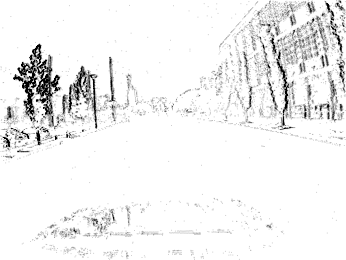}}
		&\includegraphics[clip,trim={.3cm 1.1cm .3cm 1cm},width=\linewidth]{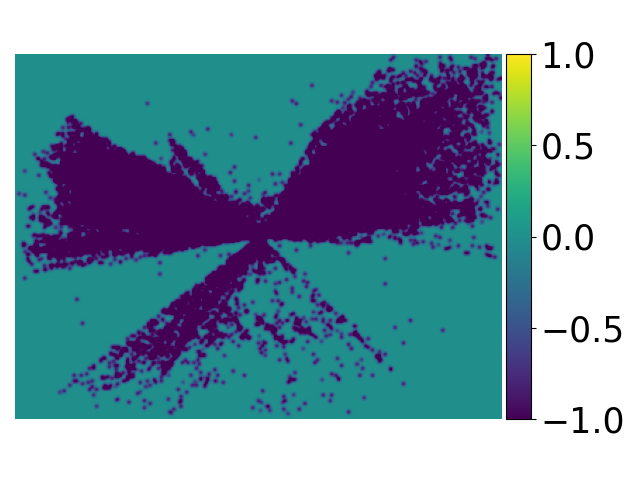}
		&\includegraphics[clip,trim={.3cm .9cm .3cm .8cm},width=\linewidth]{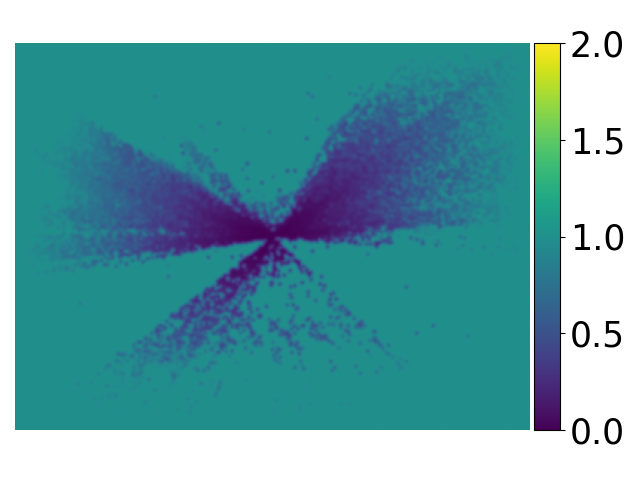}
		\\

		\multirow{2}{*}{\rotatebox{90}{\makecell{DSEC~~\cite{Gehrig21ral}}}}
		&\gframe{\includegraphics[width=\linewidth]{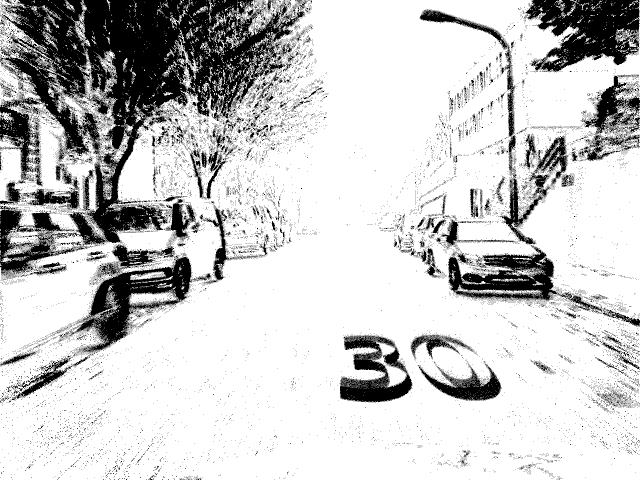}}
		&\gframe{\includegraphics[width=\linewidth]{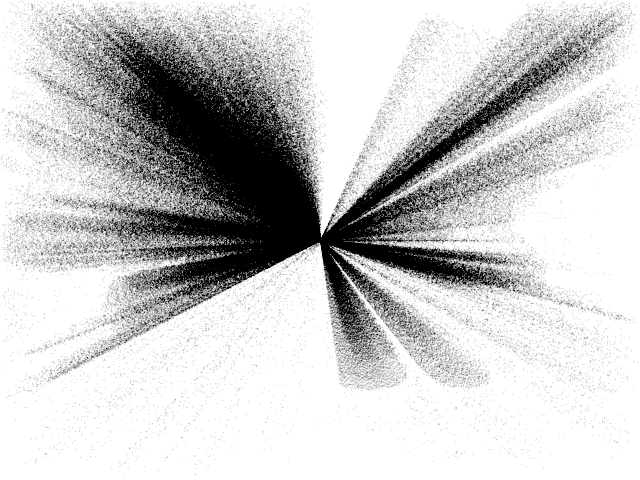}}
		&\gframe{\includegraphics[width=\linewidth]{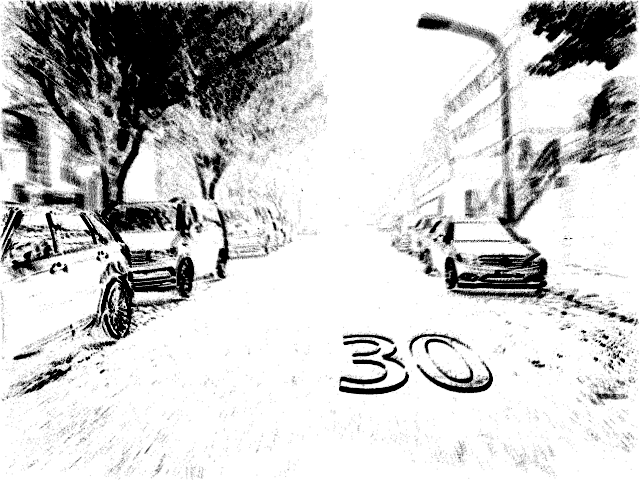}}
		&\includegraphics[clip,trim={.3cm 1.1cm .3cm 1cm},width=\linewidth]{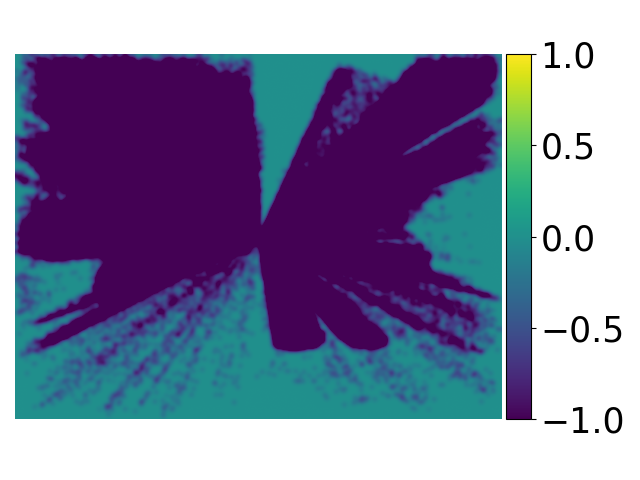}
		&\includegraphics[clip,trim={.3cm .9cm .3cm .8cm},width=\linewidth]{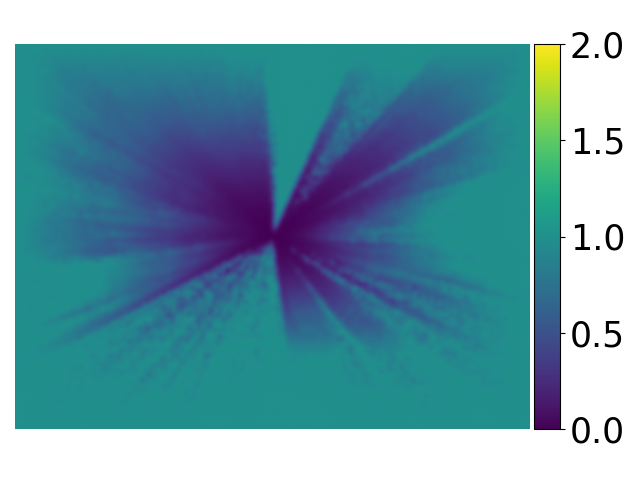}
		\\

		&\gframe{\includegraphics[width=\linewidth]{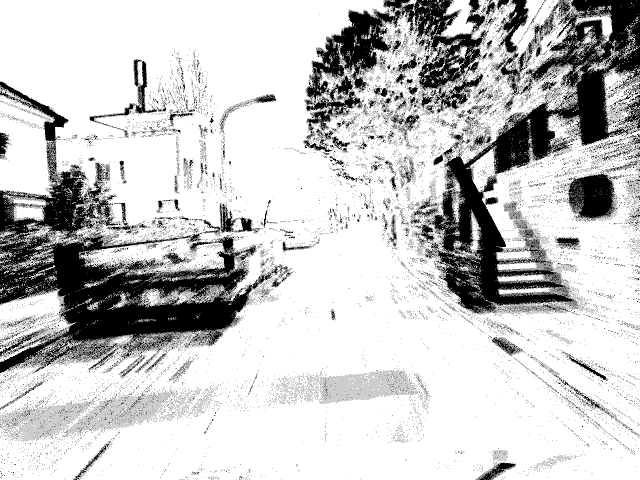}}
		&\gframe{\includegraphics[width=\linewidth]{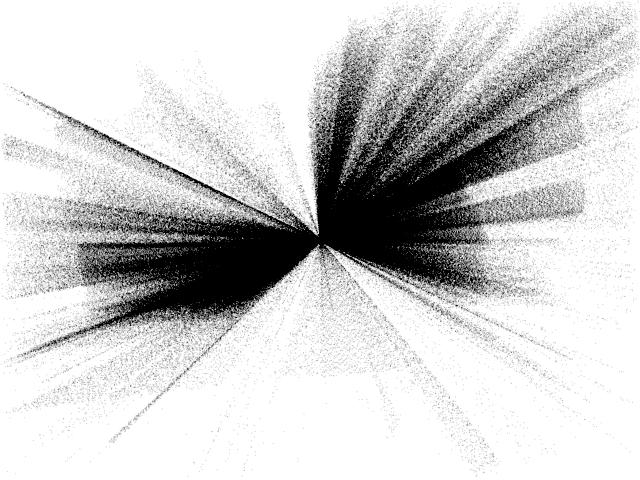}}
		&\gframe{\includegraphics[width=\linewidth]{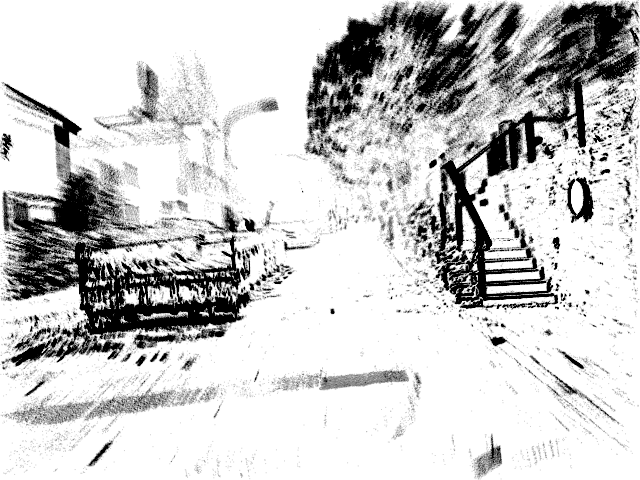}}
		&\includegraphics[clip,trim={.3cm 1.1cm .3cm 1cm},width=\linewidth]{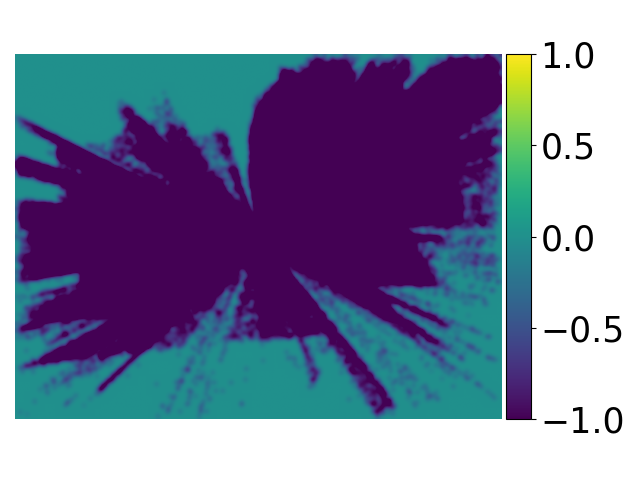}
		&\includegraphics[clip,trim={.3cm .9cm .3cm .8cm},width=\linewidth]{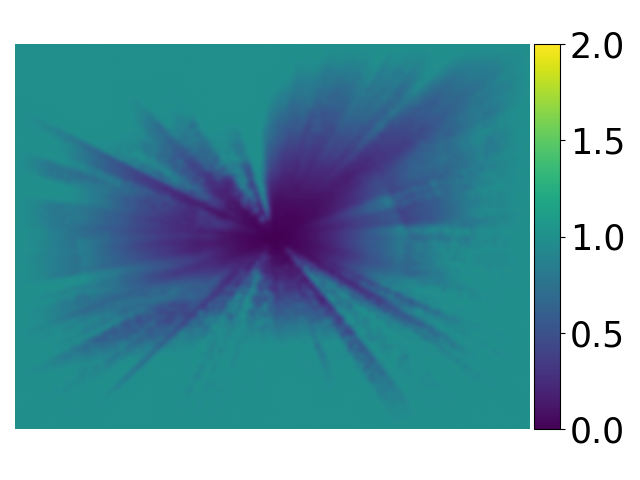}
		\\

		\multirow{2}{*}{\rotatebox{90}{\makecell{boxes\_rot~~\cite{Mueggler17ijrr}}}}
		&\gframe{\includegraphics[width=\linewidth]{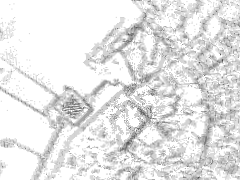}}
		&\gframe{\includegraphics[width=\linewidth]{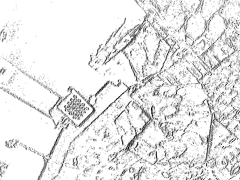}}
		&\gframe{\includegraphics[width=\linewidth]{images/fig_large/boxes/pred_warp2658.png}}
		&\includegraphics[clip,trim={.3cm 1.1cm .3cm 1cm},width=\linewidth]{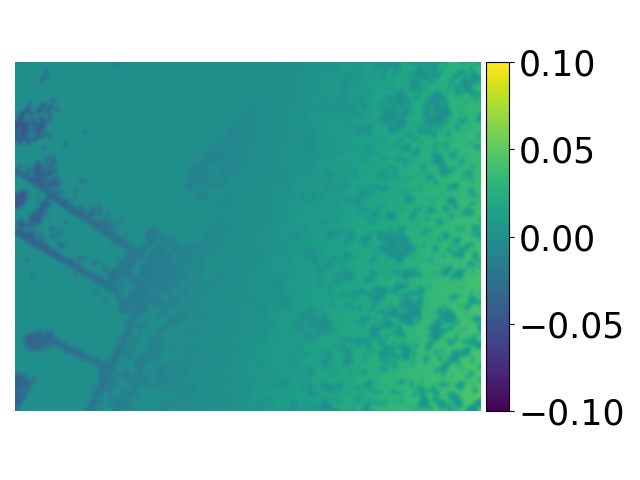}
		&\includegraphics[clip,trim={.3cm 1.1cm .3cm 1cm},width=\linewidth]{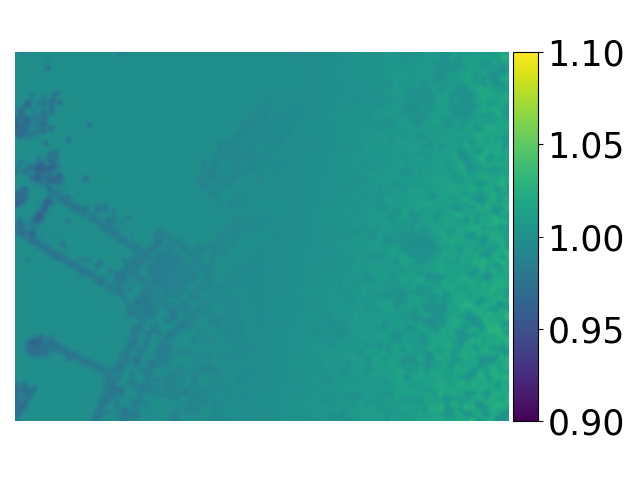}
		\\

		&\gframe{\includegraphics[width=\linewidth]{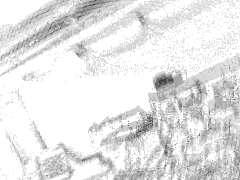}}
		&\gframe{\includegraphics[width=\linewidth]{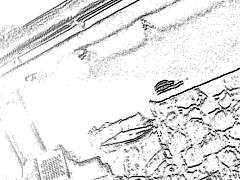}}
		&\gframe{\includegraphics[width=\linewidth]{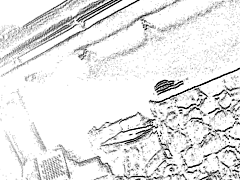}}
		&\includegraphics[clip,trim={.3cm 1.1cm .3cm 1cm},width=\linewidth]{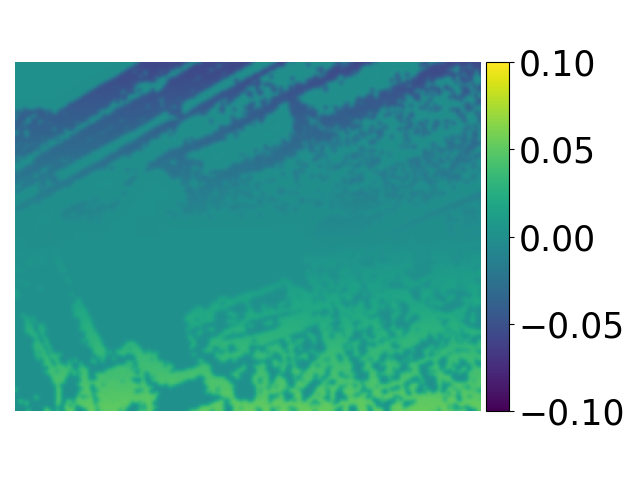}
		&\includegraphics[clip,trim={.3cm 1.1cm .3cm 1cm},width=\linewidth]{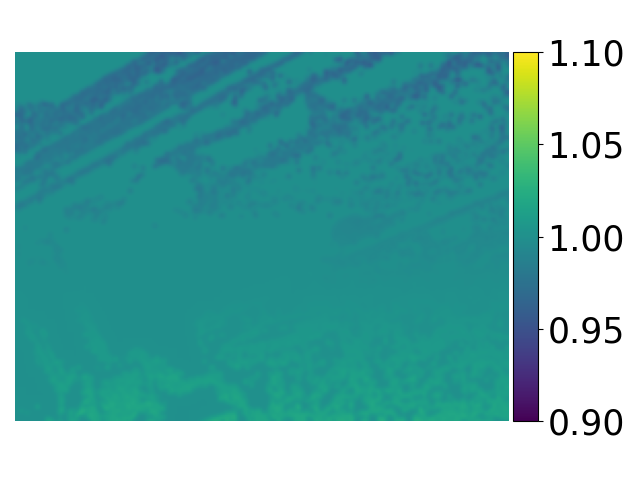}
		\\
		
		\multirow{2}{*}{\rotatebox{90}{\makecell{dynamic\_rot~~\cite{Mueggler17ijrr}}}}
		&\gframe{\includegraphics[width=\linewidth]{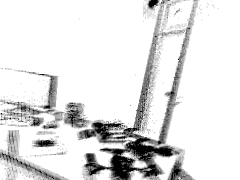}}
		&\gframe{\includegraphics[width=\linewidth]{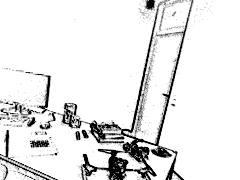}}
		&\gframe{\includegraphics[width=\linewidth]{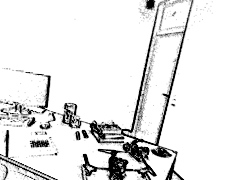}}
		&\includegraphics[clip,trim={.3cm 1.1cm .3cm 1cm},width=\linewidth]{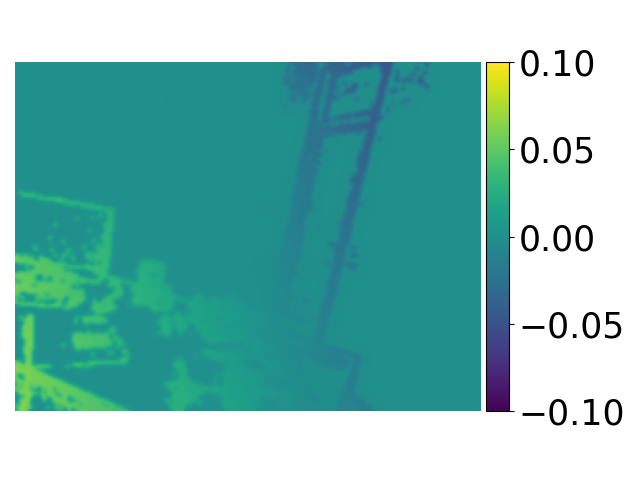}
		&\includegraphics[clip,trim={.3cm 1.1cm .3cm 1cm},width=\linewidth]{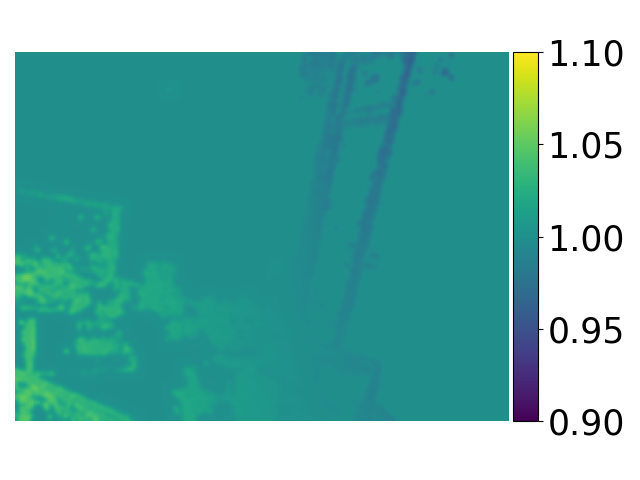}
		\\

		&\gframe{\includegraphics[width=\linewidth]{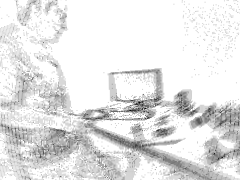}}
		&\gframe{\includegraphics[width=\linewidth]{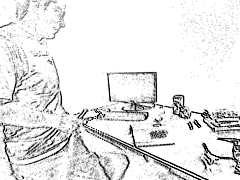}}
		&\gframe{\includegraphics[width=\linewidth]{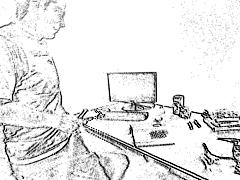}}
		&{\includegraphics[clip,trim={.3cm 1.1cm .3cm 1cm},width=\linewidth]{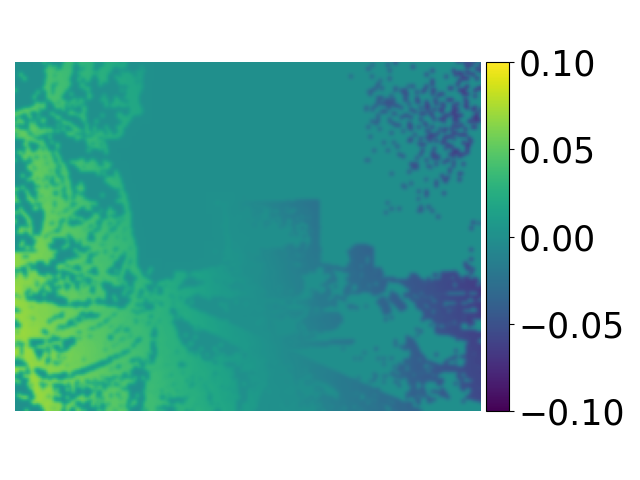}}
		&{\includegraphics[clip,trim={.3cm 1.1cm .3cm 1cm},width=\linewidth]{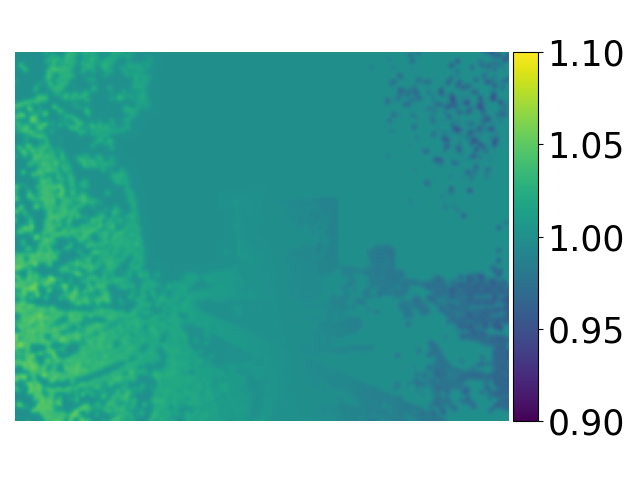}}
		\\

		& (\textbf{a})
		& (\textbf{b})
		& (\textbf{c})
		& (\textbf{d})
		& (\textbf{e})
		\\
	\end{tabular}
	}
	\caption{\emph{Proposed regularizers and collapse analysis}.
	The scene motion is approximated by 1-DOF warp (zoom in/out) for MVSEC \cite{Zhu18ral} and DSEC \cite{Gehrig21ral} sequences, and 3-DOF warp (rotation) for boxes and dynamic ECD sequences \cite{Mueggler17ijrr}.
	(\textbf{a}) Original events. %
	(\textbf{b}) Best warp without regularization. Event collapse happens for 1-DOF warp.
	(\textbf{c}) Best warp with regularization.
	(\textbf{d}) Divergence map (\eqref{eq:DIWE} is zero-based).
	(\textbf{e}) Deformation map (\eqref{eq:iwa}, centered at 1).
	Our regularizers successfully penalize event collapse and do not damage non-collapsing scenarios. 
	}
	\label{fig:main_compare}
\end{figure}

By contrast, our regularizers (Divergence and Deformation rows) work well to mitigate the collapse, as observed in smaller AEE and $N$PE values.
Compared with the values of no regularizer or whitening \cite{Nunes21pami}, our regularizers achieve more than 90\% improvement for AEE on average.
The AEE values are high for optical flow standards ($4\mbox{--}8$ pix in MVSEC vs.~$0.5\mbox{--}1$ pixel \cite{Zhu19cvpr}, or $10\mbox{--}20$ pix in DSEC vs.~$2\mbox{--}5$ pix \cite{Gehrig21threedv}); however, this is due to the fact that the warps used have very few DOFs ($\leq$4) compared to the considerably higher DOFs ($2N_p$) of optical flow estimation algorithms.
The same reason explains the high 3PE values (standard in \cite{Geiger13ijrr}): using an end-point error threshold of 3 pix to consider that the flow is correctly estimated does not convey the intended goal of inlier/outlier classification for the low-DOF warps used. 
This is the reason why Tables \ref{tab:main_mvsec} and \ref{tab:main_dsec} also report 10PE, 20PE metrics, and the values for the identity warp (zero flow). 
As expected, for the range of AEE values in the tables, the 10PE and 20PE figures demonstrate the large difference between methods suffering from collapse (20PE $>$ 80\%) and those that do not (20PE $<$ 1.1\% for MVSEC and $<$22.6\% for DSEC).

The FWL values of our regularizers are moderately high ($\geq$1), indicating that event alignment is better than that of the identity warp.
However, because the FWL depends on the number of events \cite{Stoffregen20eccv}, it is not easy to establish a global threshold to classify each method as suffering from collapse or not. The AEE, 10PE, and 20PE are better for such a~classification.

Tables \ref{tab:main_mvsec} and \ref{tab:main_dsec} also include the results of the use of both regularizers simultaneously (``Div. + Def.'').
The results improve across all sequences if the data fidelity term is given by the variance loss, whereas they remain approximately the same for the gradient magnitude loss.
Regardless of the choice of the proposed regularizer, the results in these tables clearly show the effectiveness of our proposal, i.e., the large improvements compared with prior works (rows ``No regularizer'' and \cite{Nunes21pami}).

The collapse results are more visible in Figure~\ref{fig:main_compare}, where we used the variance loss.
Without a regularizer, the events collapse in the MVSEC and DSEC sequences.
Our regularizers successfully mitigate overfitting, having a remarkable impact on the estimated~motion.

\subsection{Effect of the Regularizers on Well-Posed Warps}
\label{sec:experim:rotation}

Table \ref{tab:main_3dof} shows the results on the ECD dataset for a well-posed warp (3-DOF rotational motion, in the benchmark).
We use the variance loss and the Adam optimizer \cite{Kingma15iclr} with 100~iterations.
All values in the table (RMS error and FWL, with and without regularization, are very similar, 
indicating that: 
(i) our regularizers do not affect the motion estimation algorithm, and
(ii) results without regularization are good due to the well-posed warp.
This is qualitatively shown in the bottom part of Figure \ref{fig:main_compare}.
The fluctuations of the divergence and deformation values away from those of the identity warp ($0$ and $1$, respectively) are at least one order of magnitude smaller than the collapse-enabled warps (e.g., $0.2$ vs.~$2$).

\begin{table}[H]
\centering
\caption{Results on ECD dataset \cite{Mueggler17ijrr}.
}
\sisetup{round-mode=places,round-precision=3}
\label{tab:main_3dof}
\setlength{\cellWidtha}{\textwidth/5-2\tabcolsep+0.4in}
\setlength{\cellWidthb}{\textwidth/5-2\tabcolsep-0.1in}
\setlength{\cellWidthc}{\textwidth/5-2\tabcolsep-0.1in}
\setlength{\cellWidthd}{\textwidth/5-2\tabcolsep-0.1in}
\setlength{\cellWidthe}{\textwidth/5-2\tabcolsep-0.1in}
\scalebox{.8}[.8]{
\setlength{\tabcolsep}{2pt}
\begin{tabular}{l*{4}{S[table-format=3.4]}}
\toprule
 & \multicolumn{2}{c}{\textbf{boxes\_rot}}
 & \multicolumn{2}{c}{\textbf{dynamic\_rot}}
 \\
 \cmidrule(l{1mm}r{1mm}){2-5}
&\text{\textbf{RMS} $\boldsymbol{\downarrow}$} & \text{\textbf{FWL} $\boldsymbol{\uparrow}$}
&\text{\textbf{RMS} $\boldsymbol{\downarrow}$} & \text{\textbf{FWL} $\boldsymbol{\uparrow}$}
\\
\midrule 
Ground truth pose & \_ & 1.55901691 & \_  & 1.414303624 \\
No regularizer & 8.8578556 & 1.561590243 & 4.823395392  & 1.420484652  \\
Divergence  (Ours) & 9.236853563 & 1.558060096 &  4.826241768 & 1.420484639  \\
Deformation  (Ours) & 8.664275512 & 1.560753596 & 4.822492508  & 1.4204841  \\

\bottomrule
\end{tabular}
}
\end{table}

\subsection{Sensitivity Analysis}
\label{sec:experim:sensitivity}

The landscapes of loss functions as well as sensitivity analysis of $\lambda$ are shown in Figure~\ref{fig:objfunc:compare}, for the MVSEC experiments. 
Without regularizer ($\lambda = 0$), all objective functions tested (variance, gradient magnitude, and average timestamp \cite{Zhu19cvpr}) suffer from event collapse, which is the undesired global minimum of \eqref{eq:compositeObjective}.
Reaching the desired local optimum depends on the optimizing algorithm and its initialization
(e.g., starting gradient descent close enough to the local optimum).
Our regularizers (divergence and deformation) change the landscape:
the previously undesired global minimum becomes local, and the desired minimum becomes the new global one as $\lambda$ increases.

Specifically, the larger the weight $\lambda$, the smaller the effect of the undesired minimum (at $h_z = 1$). 
However, this is true only within some reasonable range: a too large $\lambda$ discards the data-fidelity part $G$ in \eqref{eq:compositeObjective}, which is unwanted because it would remove the desired local optimum (near $h_z\approx 0$). 
Minimizing \eqref{eq:compositeObjective} with only the regularizer is not sensible.

Observe that for completeness, we include the average timestamp loss in the last column.
However, this loss also suffers from an undesired optimum in the expansion region ($h_z \approx -1$).
Our regularizers could be modified to also remove this undesired optimum,
but investigating this particular loss, which was proposed as an alternative to the original contrast loss, is outside the scope of this work.

\global\long\def\figWidth{0.31\linewidth}
\begin{figure}[t]
	\centering
    {\scriptsize
    \setlength{\tabcolsep}{2pt}
	\begin{tabular}{
	>{\centering\arraybackslash}m{0.3cm} 
	>{\centering\arraybackslash}m{\figWidth} 
	>{\centering\arraybackslash}m{\figWidth} 
	>{\centering\arraybackslash}m{\figWidth}}
		\\

        \rotatebox{90}{\makecell{No regularizer}}
		&\includegraphics[clip,trim={0.3cm 0.4cm 0.3cm 0},width=\linewidth]{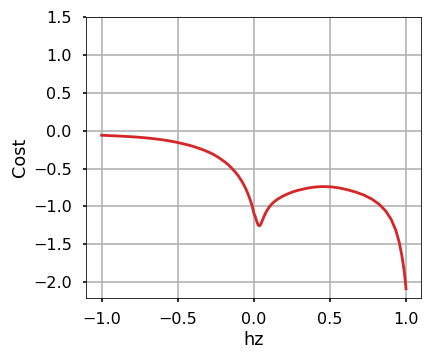}
		&\includegraphics[clip,trim={0.3cm 0.4cm 0.3cm 0},width=\linewidth]{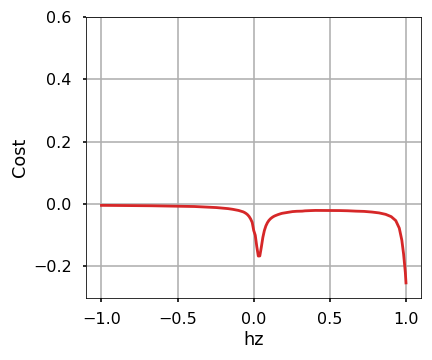}
		&\includegraphics[clip,trim={0.3cm 0.4cm 0.3cm 0},width=\linewidth]{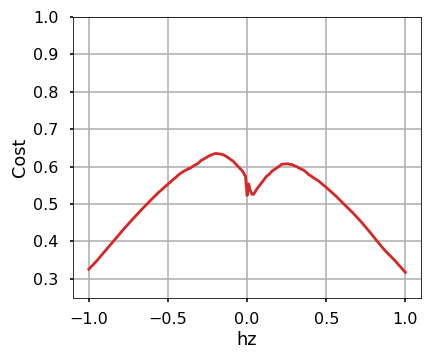}
		\\

        \rotatebox{90}{\makecell{Divergence}}
		&\includegraphics[clip,trim={0.3cm 0.4cm 0.3cm 0},width=\linewidth]{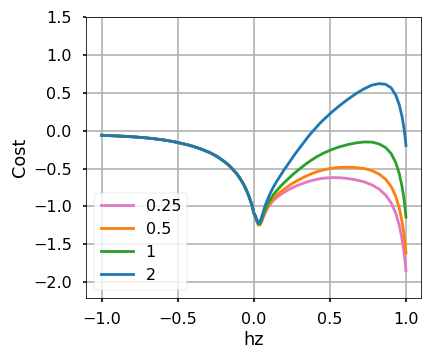}
		&\includegraphics[clip,trim={0.3cm 0.4cm 0.3cm 0},width=\linewidth]{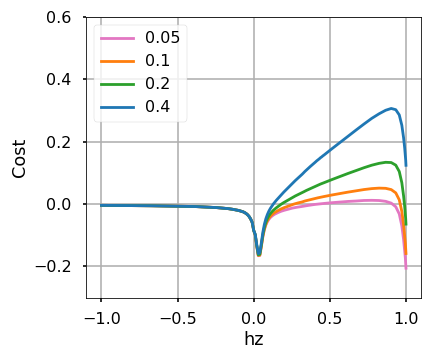}
		&\includegraphics[clip,trim={0.3cm 0.4cm 0.3cm 0},width=\linewidth]{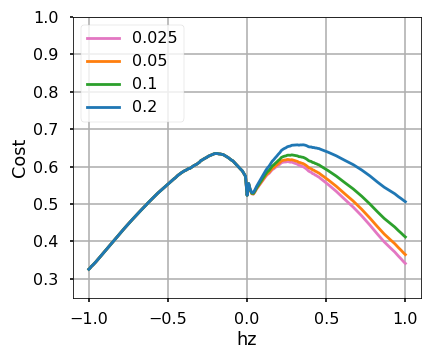}
		\\

        \rotatebox{90}{\makecell{Deformation}}
		&\includegraphics[clip,trim={0.3cm 0.4cm 0.3cm 0},width=\linewidth]{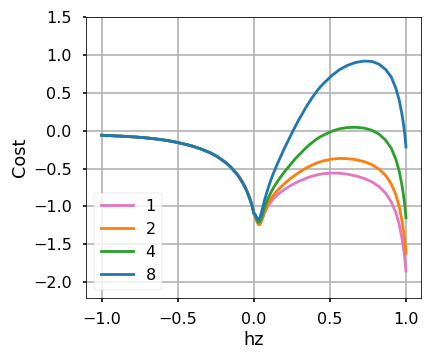}
		&\includegraphics[clip,trim={0.3cm 0.4cm 0.3cm 0},width=\linewidth]{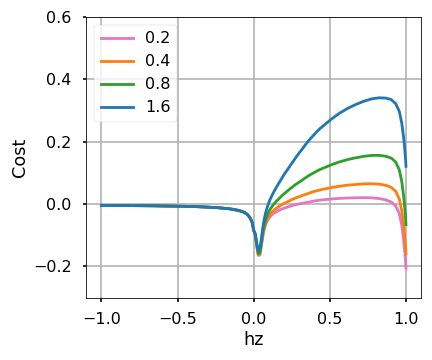}
		&\includegraphics[clip,trim={0.3cm 0.4cm 0.3cm 0},width=\linewidth]{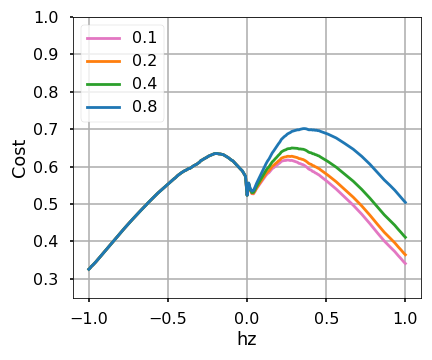}
		\\

		& (\textbf{a}) 
		& (\textbf{b})
		& (\textbf{c})
		\\
	\end{tabular}
	}
	\caption{\emph{Cost function landscapes over the warp parameter $h_z$ for}: 
	(\textbf{a}) Image variance \cite{Gallego18cvpr}, 
	(\textbf{b}) gradient magnitude \cite{Gallego19cvpr}, 
	and (\textbf{c}) mean square of average timestamp \cite{Zhu19cvpr}. 
	Data from MVSEC \cite{Zhu18ral} with dominant forward motion.
	The legend weights denote $\lambda$ in \eqref{eq:compositeObjective}.
	}
	\label{fig:objfunc:compare}
\end{figure}

\subsection{Computational Complexity}
Computing the regularizer(s) requires more computation than the non-regularized objective. 
However, complexity is linear with the number of events and the number of pixels, which is an advantage, and the warped events are reutilized to compute the DIWE or IWA. 
Hence, the runtime is less than doubled (warping is the dominant runtime term~\cite{Gallego19cvpr} and is computed only once).
The computational complexity of our regularized CMax framework is $O(\numEvents + \numPixels)$, the same as that of the non-regularized one.

\subsection{Application to Motion Segmentation}

Although most of the results on standard datasets comprise stationary scenes, 
we have also provided results on a dynamic scene (from dataset~\cite{Mueggler17ijrr}). 
Because the time spanned by each set of events processed is small, the scene motion is also small (even for complicated objects like the person in the bottom row of Figure \ref{fig:main_compare}), hence often a single warp fits the scene reasonably well. 
In some scenarios, a single warp may not be enough to fit the event data because there are distinctive motions in the scene of equal importance. 
Our proposed regularizers can be extended to such more complex scene motions.
To this end, we demonstrate it with an example in Figure \ref{fig:motion_seg}.

\def\figWidth{0.333\linewidth} \def\figWidthLong{0.43\linewidth}

\begin{figure}[H]
	\centering
    {\scriptsize
    \setlength{\tabcolsep}{2pt}
	\begin{tabular}{
	>{\centering\arraybackslash}m{0.3cm}
	>{\centering\arraybackslash}m{\figWidth} 
	>{\centering\arraybackslash}m{\figWidthLong}
	}
		& IWE with segmentation
		& Divergence map\;\;\;\;\;\\
	    
		{\rotatebox{90}{\makecell{Without regularizer}}}
		&\gframe{\includegraphics[width=\linewidth]{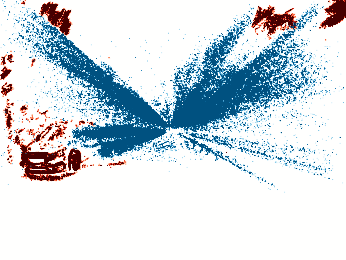}}
		&\includegraphics[clip,trim={.3cm 1.1cm .3cm 1cm},width=\linewidth]{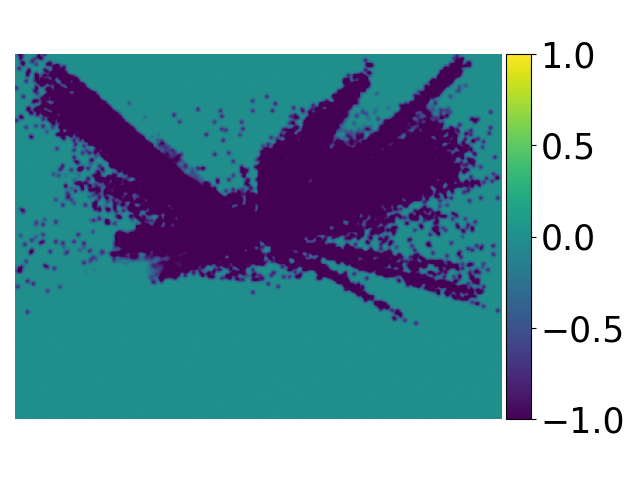}
        \\\addlinespace[.5ex]

		{\rotatebox{90}{\makecell{With regularizer~~(Ours)}}}
		&\gframe{\includegraphics[width=\linewidth]{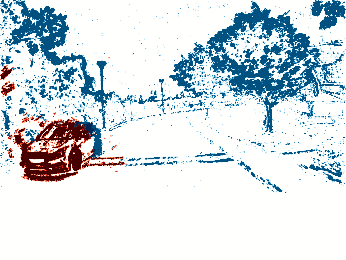}}
		&\includegraphics[clip,trim={.3cm 1.1cm .3cm 1cm},width=\linewidth]{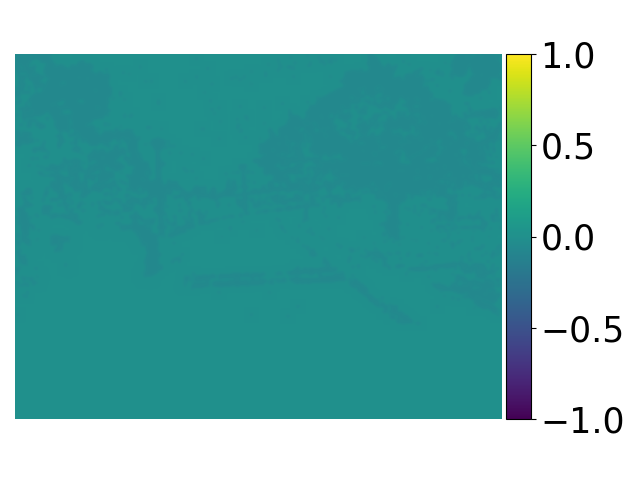}
		\\

		& (\textbf{a})
		& (\textbf{b})
		\\
	\end{tabular}
	}
	\caption{{\emph{Application to Motion Segmentation}. 
	(\textbf{a}) Output IWE, whose colors (red and blue) represent different clusters of events (segmented according to motion).
	(\textbf{b}) Divergence map. The range of divergence values is larger in the presence of event collapse than in its absence.
	Our regularizer (divergence in this example) mitigates the event collapse for this complex motion, even with an independently moving object (IMO) in the scene.
	}
	}
	\label{fig:motion_seg}
\end{figure}

Specifically, we use the MVSEC dataset, in a clip where the scene consists of two motions: 
the ego-motion (forward motion of the recording vehicle) and the motion of a car driving in the opposite direction in a nearby lane (an independently moving object---IMO).
We model the scene by using the combination of two warps. 
Intuitively, the 1-DOF warp \eqref{eq:warp:hz} describes the ego-motion, while the feature flow (2 DOF) describes the IMO.
Then, we apply the contrast maximization approach (augmented with our regularizing terms) and the expectation-maximization scheme in \cite{Stoffregen19iccv} to segment the scene, to determine which events belong to each motion.
The results in Figure \ref{fig:motion_seg} clearly show the effectiveness of our regularizer, even for such a commonplace and complex scene. 
Without regularizers, (i) event collapse appears in the ego-motion cluster of events and (ii) a considerable portion of the events that correspond to ego-motion are assigned to the second cluster (2-DOF warp), thus causing a segmentation failure.
Our regularization approach mitigates event collapse (bottom row of Figure \ref{fig:motion_seg}) and provides the correct segmentation: the 1-DOF warp fits the ego-motion and the feature flow (2-DOF warp) fits the IMO.

\section{Conclusions}
\label{sec:conclusion}

We have analyzed the event collapse phenomenon of the CMax framework and proposed collapse metrics using first principles of space-time deformation, inspired by differential geometry and physics.
Our experimental results on publicly available datasets demonstrate that the proposed divergence and area-based metrics mitigate the phenomenon for collapse-enabled warps and do not harm well-posed warps.
To the best of our knowledge, our regularizers are the only effective solution compared to the unregularized CMax framework and whitening.
Our regularizers achieve, on average, more than 90\% improvement on optical flow endpoint error calculation (AEE) on collapse-enabled warps.

This is the first work that focuses on the paramount phenomenon of event collapse.
No prior work has analyzed this phenomenon in such detail or proposed new regularizers without additional data or reparameterizing the search space \cite{Zhu19cvpr,Nunes21pami,Peng21pami}. 
As we analyzed various warps from 1 DOF to 4 DOFs,
we hope that the ideas presented here inspire further research to tackle more complex warp models.
Our work shows how the divergence and area-based deformation can be computed for warps given by analytical formulas.
For more complex warps, like those used in dense optical flow estimation \cite{Zhu19cvpr,Paredes21neurips}, 
the divergence or area-based deformation could be approximated by using finite difference formulas.

\vspace{6pt}

%

\appendixtitles{yes} %
\appendixstart
\appendix

\section{Warp Models, Jacobians and Flow Divergence}
\label{sec:warpmodels}

\subsection{\textls[-25]{Planar Motion --- Euclidean Transformation on the Image Plane, $SE(2)$
}}
\label{sec:warpmodels:planarmotion}
If the point trajectories of an isometry are $\bx(t)$, the warp is given by \cite{Nunes21pami}
\begin{equation}
\begin{pmatrix}
\bx'_k\\1
\end{pmatrix}
\sim 
\begin{pmatrix}
\Rot(t_k\omega_Z) & t_k \velflow\\
\bzero^\top & 1
\end{pmatrix}^{-1}
\begin{pmatrix}
\bx_k\\1
\end{pmatrix},
\end{equation}
where $\velflow, \omega_Z$ comprise the 3 DOFs of a translation and an in-plane rotation.
The in-plane rotation is 
\begin{equation}
\Rot(\phi) = \begin{pmatrix}
\cos \phi & \;- \sin \phi\\ \sin \phi & \cos \phi
\end{pmatrix} .   
\end{equation}

Since
\begin{equation}
\label{eq:invmatrixtwoblock}
\begin{pmatrix}
\mA & \bvec\\
\bzero^\top & 1
\end{pmatrix}^{-1}
=\begin{pmatrix}
\mA^{-1} & -\mA^{-1} \bvec\\
\bzero^\top & 1
\end{pmatrix}
\end{equation}
and $\Rot^{-1}(\phi) = \Rot(-\phi)$, we have
\begin{equation}
\begin{pmatrix}
\bx'_k\\1
\end{pmatrix}
\sim 
\begin{pmatrix}
\Rot(-t_k\omega_Z) & - \Rot(-t_k\omega_Z) (t_k \velflow)\\
\bzero^\top & 1
\end{pmatrix}
\begin{pmatrix}
\bx_k\\1
\end{pmatrix}.
\end{equation}

Hence, in Euclidean coordinates the warp is
\begin{equation}
\label{eq:warpIsometry}
\bx'_k = \Rot(-t_k\omega_Z) (\bx_k - t_k \velflow).
\end{equation}

The Jacobian and its determinant are:
\begin{equation}
\mJ_k = \prtl{\bx'_k}{\bx_k} = \Rot(-t_k\omega_Z),
\end{equation}
\begin{equation}
\det(\mJ_k) = 1.
\end{equation}

The flow corresponding to \eqref{eq:warpIsometry} is:
\begin{equation}
\flow = \prtl{\bx'}{t} = \Rot^\top\Bigl(\frac{\pi}{2}+t\omega_Z\Bigr)(\bx - t \velflow)\omega_Z - \Rot(-t\omega_Z)\velflow,
\end{equation}
whose divergence is
\begin{equation}
\nabla \cdot \flow = -2\omega_Z \sin(t\omega_Z).
\end{equation}

Hence, for small angles $|t\omega_Z|\ll 1$, the divergence of the flow vanishes.

In short, this warp has the same determinant and approximate zero divergence as the 2-DOF feature flow warp (Section \ref{sec:method:featureflow}), which is well-behaved.
Note, however, that the trajectories are not straight in space-time.

\subsection{3-DOF Camera Rotation, $SO(3)$}
\label{sec:warpmodels:rotationcamera}
Using calibrated and homogeneous coordinates, the warp is given by \cite{Gallego17ral,Gallego18cvpr}
\begin{equation}
\bx_k^{h\prime} \sim \Rot(t_k \angvel)\,\bx^h_k,
\end{equation}
where $\bparams = \angvel = (\omega_{1},\omega_{2},\omega_{3})^{\top}$ is the angular velocity, 
and $\Rot$ ($3\times 3$ rotation matrix in space) is parametrized using exponential coordinates (Rodrigues rotation formula \cite{Murray94book,Gallego14jmiv}).

By the chain rule, the Jacobian is:
\begin{equation}
    \mJ_k = \prtl{\bx'_k}{\bx_k} 
    = \prtl{\bx'_k}{\bx^{h\prime}_k} \prtl{\bx^{h\prime}_k}{\bx^h_k} \prtl{\bx^h_k}{\bx_k}
    = \frac{1}{(\bx^{h\prime}_k)_3}\begin{pmatrix} 1 & 0 & -x'_k\\ 0 & 1 & -y'_k \end{pmatrix} \Rot(t_k \angvel) 
    \begin{pmatrix} 1 & 0\\ 0 & 1\\ 0 & 0\end{pmatrix}.
\end{equation}

Letting $\br_{3,k}^\top$ be the third row of $\Rot(t_k \angvel)$, and using (32)--(34) in~\cite{Gallego11tgrs}, gives
\begin{equation}
\label{eq:rot:detJacArea:app}
\det(\mJ_k) = (\br_{3,k}^\top \bx^h_k)^{-3}.
\end{equation}

\subsubsection*{Connection between divergence and deformation maps}
If the rotation angle $t_k \|\angvel\|$ is small, using the first two terms of the exponential map we approximate 
$\Rot (t_k \angvel) \approx \mId + (t_k \angvel)^\wedge$, 
where the hat operator $^\wedge$ in $SO(3)$ represents the cross product matrix \cite{Barfoot15book}.
Then, $\br_{3,k}^\top \bx^h_k \approx (-t_k \omega_2, t_k \omega_1, 1)^\top (x_k,y_k,1) 
= 1 + (y_k\omega_1 - x_k\omega_2) t_k$.
Substituting this expression into \eqref{eq:rot:detJacArea:app} and using the first two terms in Taylor's expansion around $z=0$ of $(1+z)^{-3} \approx 1-3z+6z^2$ (convergent for $|z|<1$) gives
$\det(\mJ_k) \approx 1 + 3(x_k\omega_2 - y_k\omega_1) t_k$.
Notably, the divergence \eqref{eq:rot:flowdivergence} and the approximate amplification factor depend linearly on $3(x_k\omega_2 - y_k\omega_1)$. 
This resemblance is seen in the divergence and deformation maps of the bottom rows in Figure \ref{fig:main_compare} (ECD dataset).

\subsection{4-DOF In-Plane Camera Motion Approximation}
\label{sec:warpmodels:maryland}

The warp presented in \cite{Mitrokhin18iros}, 
\begin{equation}
\label{eq:MarylandForuDofWarp}
\bx'_k = \bx_k - t_k\,\bigl( \velflow + (h_z+1)\Rot(\phi)\bx_k-\bx_k \bigr)
\end{equation}
has 4 DOFs: $\btheta = (\velflow, \phi, h_z)^\top$. 
The Jacobian and its determinant are:
\begin{equation}
\mJ_k = \prtl{\bx'_k}{\bx_k} = (1+t_k) \mId - (h_z+1) t_k \Rot(\phi),
\end{equation}
\begin{equation}
\det(\mJ_k) = (1+t_k)^2 - 2 (1+t_k) t_k (h_z+1) \cos\phi +t_k^2 (h_z+1)^2.
\end{equation}

The flow corresponding to \eqref{eq:MarylandForuDofWarp} is given by 
\begin{equation}
    \flow = \prtl{\bx'}{t} = - \bigl( \velflow + (h_z+1)\Rot(\phi)\bx-\bx \bigr),
\end{equation}
whose divergence is:
\begin{align}
    \nabla \cdot \flow &= - (h_z+1)\nabla \cdot \bigl(\Rot(\phi)\bx\bigr) + \nabla \cdot \bx\\
    &= 2 - 2(h_z+1)\cos(\phi).
\end{align}

As particular cases of this warp, one can identify:
\begin{itemize}
    \item 1-DOF Zoom in/out ($\velflow=\bzero, \phi=0$).
$\bx'_k = (1 - t_k h_z) \bx_k$.

\item 2-DOF translation ($\phi=0, h_z=0$).
$\bx'_k = \bx_k - t_k \velflow$.

\item 1-DOF ``rotation'' ($\velflow=\bzero, h_z=0$).
$\bx'_k = \bx_k - t_k\,\bigl( \Rot(\phi)\bx_k-\bx_k \bigr)$.\\
Using a couple of approximations of the exponential map in $SO(2)$, we obtain
\begin{align}
\bx'_k & = \bx_k - t_k\,\bigl( \Rot(\phi)-\mId \bigr)\bx_k \\
&\approx \bx_k - t_k \phi^\wedge \bx_k & \text{if } \phi \text{ is small}\\
&= (\mId + (-t_k \phi)^\wedge) \bx_k\\
&\approx \Rot(-t_k \phi) \bx_k & \text{if } t_k\phi \text{ is small}.\label{eq:maryland:approxrot}
\end{align} 
Hence, $\phi$ plays the role of a small angular velocity $\omega_Z$ around the camera's optical axis $Z$, i.e., in-plane rotation.

\item 3-DOF planar motion (``isometry'') ($h_z=0$).
Using the previous result, the warp splits into translational and rotational components:
\begin{align}
\bx'_k & = \bx_k - t_k\,\bigl( \velflow + \Rot(\phi)\bx_k-\bx_k \bigr)\\
&\stackrel{\eqref{eq:maryland:approxrot}}{\approx} -t_k \velflow + \Rot(-t_k \phi) \bx_k.
\end{align}
\end{itemize}

\subsection{4-DOF Similarity Transformation on the Image Plane, Sim(2)}
\label{sec:warpmodels:similarity}

Another 4-DOF warp is proposed in \cite{Nunes21pami}. 
Its DOFs are the linear, angular and scaling velocities on the image plane: $\btheta = (\velflow, \omega_Z, s)^\top$.

Letting $\beta_k = 1+t_k s$, the warp is:
\begin{equation}
\begin{pmatrix}
\bx'_k\\1
\end{pmatrix}
\sim 
\begin{pmatrix}
\beta_k \Rot(t_k\omega_Z) & t_k \velflow\\
\bzero^\top & 1
\end{pmatrix}^{-1}
\begin{pmatrix}
\bx_k\\1
\end{pmatrix}.
\end{equation}

Using \eqref{eq:invmatrixtwoblock} gives
\begin{equation}
\begin{pmatrix}
\bx'_k\\1
\end{pmatrix}
\sim 
\begin{pmatrix}
\beta^{-1}_k \Rot(-t_k\omega_Z) &  -\beta^{-1}_k \Rot(-t_k\omega_Z) (t_k \velflow)\\
\bzero^\top & 1
\end{pmatrix}
\begin{pmatrix}
\bx_k\\1
\end{pmatrix}.
\end{equation}

Hence, in Euclidean coordinates the warp is
\begin{equation}
\label{eq:WarpedPointSimilarityTwo}
\bx'_k = \beta^{-1}_k \Rot(-t_k\omega_Z) (\bx_k - t_k \velflow).
\end{equation}

The Jacobian and its determinant are:
\begin{equation}
\mJ_k = \prtl{\bx'_k}{\bx_k} = \beta^{-1}_k \Rot(-t_k\omega_Z),
\end{equation}
\begin{equation}
\det(\mJ_k) = \beta^{-2}_k = \frac1{(1+t_k s)^2}.
\end{equation}

The following result will be useful to simplify equations.
For a 2D rotation $\Rot(\phi(t))$, it holds that:
\begin{align}
\label{eq:derivTwoDimRotMat}
\prtl{\Rot(\phi(t))}{t} = -\Rot^\top \Bigl(\frac{\pi}{2}-\phi \Bigr)\, \prtl{\phi}{t}.
\end{align}

To compute the flow of \eqref{eq:WarpedPointSimilarityTwo}, there are three time-dependent factors. 
Hence, applying the product rule we obtain three terms, and substituting \eqref{eq:derivTwoDimRotMat} (with $\phi = -t\omega_Z$) gives:
\begin{align}
    \flow_k = \Bigl(\prtl{\beta^{-1}_k}{t_k} \Rot(-t_k\omega_Z)
    + \beta^{-1}_k \omega_Z \Rot^\top\bigl(\frac{\pi}{2}+t_k\omega_Z\bigr)\Bigr) (\bx_k - t_k \velflow)
    - \beta^{-1}_k \Rot(-t_k\omega_Z) \velflow,
\end{align}
where, by the chain rule,
\begin{align}
\label{eq:prtinvbeta}
    \prtl{\beta^{-1}_k}{t_k} = -\beta^{-2}_k \prtl{\beta_k}{t_k} 
    = -\beta^{-2}_k s = -\frac{s}{(1+t_k s)^{2}}.
\end{align}

Hence, the divergence of the flow is:
\begin{align}
    \nabla\cdot\flow_k 
    & = \prtl{\beta^{-1}_k}{t_k} \nabla\cdot \Bigl(\Rot(-t_k\omega_Z) \bx_k\Bigr) 
    + \beta^{-1}_k \omega_Z \nabla\cdot \Bigl(\Rot^\top\bigl(\frac{\pi}{2}+t_k\omega_Z\bigr) \bx_k\Bigr)\\
    & = \prtl{\beta^{-1}_k}{t_k} 2 \cos(t_k\omega_Z)
    + \beta^{-1}_k \omega_Z 2\sin(-t_k\omega_Z)
\end{align}

The formulas for $SE(2)$ are obtained from the above ones with $s=0$ (i.e., $\beta_k=1$).

\begin{adjustwidth}{-\extralength}{0cm}

\reftitle{References}


\end{adjustwidth}

\end{document}